# Governance of Generative Artificial Intelligence for Companies

Johannes Schneider
University of Liechtenstein
Liechtenstein
johannes.schneider@uni.li
(Corresponding author)

Rene Abraham
University of Liechtenstein
Liechtenstein
rene.abraham@gmx.ch

Christian Meske
Ruhr University Bochum
Germany
christian.meske@ruhr-uni-bochum.de

**Abstract**

Generative Artificial Intelligence (GenAI) like ChatGPT has swiftly entered organizations without adequate governance, posing both opportunities and risks. Limited research addresses organizational governance from both technical and business perspectives. This gap is particularly relevant for international businesses, where differences in regulation, language, and business environments complicate governance. While multiple frameworks for AI governance exist, this needed diversity is lacking for GenAI. This review paper fills this gap by surveying recent literature to better understand the fundamental characteristics of GenAI and to adapt existing governance frameworks specifically to GenAI. The resulting framework delineates scope, objectives, and governance mechanisms designed to both harness business opportunities and mitigate risks associated with GenAI integration. We theorize a distinctive property of GenAI governance: its scope is endogenous to use. Unlike conventional organizational AI, for which governance is organized around a fixed artifact (e.g., model, intended use), GenAI allows users to reconfigure the artifact (e.g., its behavior and risk). Consequently, the object of governance is not fixed ex ante but is partly constituted through use.

## Introduction

Generative artificial intelligence (GenAI) has the potential to disrupt industries (McAfee et al., 2023; Wamba et al., 2025). It can generate business value in various ways, from enhancing content quality and employee expertise to augmenting customer acquisition and retention (Dencik et al., 2023), outperforming on product idea generation (Joosten et al., 2024), and (clinical) text summarization (Van Veen et al., 2024). Research indicates productivity gains around 25% when implementing GenAI (McAfee et al., 2023; S. Noy & Zhang, 2023). Leveraging the value of GenAI applications requires organizations to not only enable its potential, but to also mitigate its specific risks rigorously. This duality – enabling potential and mitigating risks – is the core objective of organizational AI governance (Mäntymäki et al., 2022; Schneider et al., 2022).

The properties of GenAI introduce unique risks and challenges that require attention when implementing GenAI within organizations. For instance, generation of artefacts such as images and texts introduces concerns including the generation of incorrect or harmful content, which can lead to erroneous decisions, reputational damage or legal issues including copyright infringements (Grynbaum & Mac, 2023). As prior to GenAI such artefacts could not be generated at a comparable scale with such ease, this is a novel phenomenon requiring adequate governance, e.g., training employees but also potentially mitigating risks through contractual agreements.

The autonomy and opacity of GenAI complicate the control of output quality, particularly illustrated by the phenomenon of hallucinations, where the AI produces false content which frequently appears trustworthy on first glance (Kalai et al., 2025). Machine learning models are known for their statistical behavior producing seemingly random errors. But GenAI raises concerns to the next level as its outputs are not simple decisions but outputs that are harder to assess such as complex texts that sound convincing despite containing factual errors. Minor changes to inputs can drastically change outputs (prompt sensitivity (Schneider et al., 2024b). GenAI yields novel security risks such as jailbreaking to bypass safety measures. In turn, governance must account for such challenges.

Another source of exemplary, unique governance challenges relates to the easy-to-access foundation models as emerging paradigm for solving a variety of tasks and even developing software (Schneider et al., 2024b). Foundation models are large-scale machine learning models pre-trained on extensive datasets. They can be adapted to a wide range of tasks and applications with no or little data and coding expertise that can also interact with other software (Schick et al., 2024). Individual employees can access tools like ChatGPT independently of organizational implementation, potentially leading to bottom-up deployment dynamics that challenge centralized AI governance approaches. These dynamics impact multinational organizations especially, where employees may use GenAI across jurisdictions with different legal, linguistic, and cultural requirements For instance, a marketing employee with the task to "Read a news article (mentioning the company) and summarize it" can prompt a GenAI model with the relevant article. The employee might improve the prompt over time, e.g., specify writing style, to optimize the AI's output. Platforms like OpenAI's Custom GPTs (OpenAI, 2023c) empower the employee to create and share her GenAI app globally at the click of a button. This example illustrates how the spread of prompt-based GenAI disrupts AI development and deployment, blurring the boundaries between software developers and end users. Due to GenAI's striking capabilities employees might have a strong incentive to use such tools, which might enhance their productivity but might not align well with organizational policies, highlighting the need for adequate governance. In a recent survey, more than 44% out of 48000 people reported using AI in ways that contravened organizational policies (Gillespie et al., 2025).

Besides the concerns indicated above, ill-governed GenAI implies numerous organizational risks including loss of productivity and quality (Simkute et al., 2024), IP leakage (Gillespie et al., 2025) and dual-use scenarios (Koplin, 2023). Even leading AI companies like OpenAI struggle to prevent undesired model behavior such as hallucinations, illustrating the difficulty of GenAI risk mitigation (OpenAI, 2025). Thus, although it initially appears straightforward for organizations to harness the value of GenAI, a closer examination reveals the complexities and associated risks involved that have not been observed prior to GenAI in a similar way, e.g., (i) bottom-up adoption across many organizational functions; (ii) rapidly evolving foundation models interacting with other applications and novel shortcomings such as hard to identify hallucinations; (iii) more complex outputs and tasks; and (iv) growing challenges for internationally operating firms related to regulation, language, and globally distributed technology providers. Given that international organizations are explicitly built into the research question, the paper should show *how governance differs across countries and multinational structures*, not only list country/regulation as antecedents. The present framework could largely be applied to a domestic company without modification, which weakens the international-business positioning.

This underscores the necessity of reevaluating existing AI governance approaches to suit the unique demands of GenAI. Organizations must develop tailored approaches to integrate GenAI effectively and responsibly with their operational workflows and frameworks. There are a number of existing frameworks of AI governance that provide a starting point (e.g., (Schneider et al., 2022), (Birkstedt et al., 2023)) but do not account for the specific properties of GenAI. Existing work on the governance of GenAI has focused mostly on the perspective of model providers (Esposito De Falco et al, 2026), policymakers and society (NATO, 2021; Sigfrids et al., 2023; Taeihagh, 2025) or on individual aspects such as guardrails (Leon, 2026), trust (Newman, 2023), risk (Maas, 2023; NIST, 2023; NIST, 2024; Emett, 2026), responsibility (Janssen, 2025), and ethics (Hagendorff, 2024; Huang et al., 2025; Mäntymäki et al., 2022) without considering the implementation of organizational governance holistically. Considering the unique risks and challenges described above, we highlight the imperative to reassess organizational AI governance in the context of GenAI. Consequently, we propose the following research question:

*How can the specific properties of GenAI be captured by a governance framework for organizations?*

To contribute to respective discourse with gap-filling conceptualization, we conduct a specific theorizing review (Leidner, 2018) to develop a framework for GenAI governance building on prior works (Schneider et al., 2022). That is, we identify gaps in existing frameworks (e.g., (NIST 2023; NIST, 2024; Schneider et al., 2022)) with the goal to close them based on existing knowledge identified in relevant literature. A literature review is suitable as it provides a broad perspective and there is a large number of publications on GenAI – both of which is needed for a framework. However, our systematic review of literature (Okoli & Schabram, 2015) on the governance of GenAI from a business perspective also highlights that there exist a number of concerns that seem to be relevant but were not included within the search. As such, we deemed a systematic literature review as too rigid and insufficient. In turn, we extend the taxonomy development processes described in (Nickerson et al., 2013). We adopt their process to include a conceptual-to-conceptual iteration (c2c), i.e., a comparison of existing AI governance frameworks and an assessment of the unique properties of GenAI to refine the initial concepts from the systematic literature review and existing AI frameworks in a narrative manner. The final framework describes characteristics of GenAI governance along the five dimensions of antecedents, scope, governance mechanisms, targets, and consequences. Our central theoretical contribution is to conceptualize GenAI governance scope as endogenous to use: the governed object is not fully defined ex ante but can be reconfigured through its organizational use.

## Methodology

We conduct a literature review with the goal of developing a GenAI-specific governance framework for organizations by identifying (i) organizational governance concerns that arise in light of GenAI and (ii) adapting existing frameworks for (Gen)AI governance. Our resulting framework fills gaps identified during the literature review. It resembles a specific theorizing review (Leidner, 2018) in which literature may be used to inform the gap-filling conceptualization. We began by establishing an initial corpus of articles using keywords. We also considered existing pre-GenAI frameworks with different emphasis as a foundation (e.g., (NIST, 2023; NIST, 2024; Schneider et al., 2022)). We then expanded the corpus using iterative, concept-driven search in a narrative manner to cover constructs insufficiently covered in the initial corpus. This approach balances search transparency and the ability to achieve conceptual coverage.

Our review process started with a search covering keywords (“governance” AND ("generative AI" OR "generative artificial intelligence" OR “LLMs”)) on AISnet and Google Scholar as they include recent articles and enable broad coverage, exceeding databases such as Web of Science and Scopus (Martín-Martín et al., 2021). We decided to include articles from arXiv.org as the platform serves as an important repository for current research findings in AI (Lin et al., 2020). As such, we put the burden of reviewing on us and removed articles that (i) did not fit the scope. This included many articles covering generic governance concerns rather than concerns specific to GenAI, specifically addressing legal questions targeted at policymakers, and articles concerning other fields than business. Furthermore, we removed publications that (ii) did not meet quality criteria. This included articles with a lack of references, presentations below the quality standards of typical scientific venues, or short texts such as editorials. As part of the quality appraisal, we gave preference to papers presenting evidence and only sparsely included opinion papers from which numerous were published within 3-4 months after the release of ChatGPT. We restricted the search to the years after 2021 to August 2026. In 2020 OpenAI’s GPT-3 (Brown et al., 2020) offered an early version of AI with prompt-based generative capabilities, while GenAI emerged at scale for businesses only in 2022 with the release of ChatGPT 3.5.

In total, for our keywords aiming to find an initial seed of papers Google scholar reported 15600 hits and AISNet 119. We assessed the first 2000 on Scholar (1000 prior to 2025 and 1000 only in 2025) to give us an initial seed of 58 articles. Mostly, articles talked about GenAI and governance but were not about companies although they did contain keywords such as “organization”, “company”, or “business”. Nevertheless, the initial seed contained a number of highly relevant frameworks that served as a good starting point to be further enhanced. We analyze and synthesize the insights of the initial review using the taxonomy development process described in Nickerson et al. (2013) which presents a method for creating artifacts such as typologies, taxonomies, and frameworks. The resulting artifact describes the object under study in terms of dimensions consisting of groups of characteristics. We enhanced this process to include a conceptual-to-conceptual mapping allowing us to come up with a taxonomy that leverages prior conceptualizations, which in our case are prior frameworks of GenAI Governance as well as conceptualization of GenAI that are likely impacting governance. That is, our approach extends (Nickerson et al., 2013) to include prior conceptualization by mapping them against the developed taxonomy. The enhanced process is shown in Figure 1.

The process begins with the definition of a meta-characteristic which informs the subsequent choice of other characteristics and should thus be based on the purpose of the framework. Given our research objective of describing the scope of GenAI governance approaches, our meta-characteristics are the properties of GenAI that require governance responses. Second, a researcher must identify prior conceptualizations that are used to map against the developed taxonomy. These can be taxonomies that are supposed to be enhanced or altered, e.g., as the field has progressed, and novel knowledge is available. In our case, we focus on prior AI frameworks, i.e., (Schneider et al., 2022) and (NIST, 2023) with its GenAI extension (NIST, 2024). It can also consist of conceptualizations that directly impact the developed taxonomy and should be addressed during the development process. In our case, we aim to include key concepts of GenAI as outlined in the introduction that are likely to impact governance. Third, the researcher must define ending conditions to determine when to terminate their iterative revision of the framework. We use the ending conditions suggested in the original work by (Nickerson et al., 2013). Nickerson et al propose two approaches to iteratively identifying dimensions and characteristics for taxonomy development until the ending conditions are met. The identification is done either based on an empirical-to-conceptual (e2c) or conceptual-to-empirical (c2e) approach. We add a conceptual-to-(prior) conceptual (c2c) iteration. This iteration relies on a set of external, existing concepts and maps them against identified dimensions and characteristics shown to map between the prior conceptualization and the (novel) taxonomy with the goal to identify gaps and reassess dimensions and characteristics.

Specifically, we begin with an e2c step by deductively deriving an initial conceptualization of characteristics and dimensions from the reviewed literature. We thereby focus on aspects of how organizations can enable opportunities while mitigating the risks of GenAI. We select an initial seed of relevant literature using keyword search as detailed above. We then perform a c2c step to further refine our initial set of characteristics and dimensions using two ways of mapping:

(1) Framework comparison: we compare our findings against existing frameworks specifically on AI governance (Schneider et al., 2022) and risk (NIST, 2023; NIST, 2024). The comparison consists of (a) mapping dimensions and characteristics to identify what might be missing or novel in our framework, and (b) assessing the grouping of characteristics into dimensions, i.e., is the structure of the existing framework appropriate or would a novel structure be more suitable.

(2) Mapping of the initial dimensions and characteristics to the unique properties of GenAI. We review additional literature to inform our understanding of respectively relevant properties discussed in the introduction. As outcome of the c2c iteration we arrive at a set of potential concepts that lack empirical support in our context as well as an adjusted framework, e.g., regrouped dimensions and characteristics.

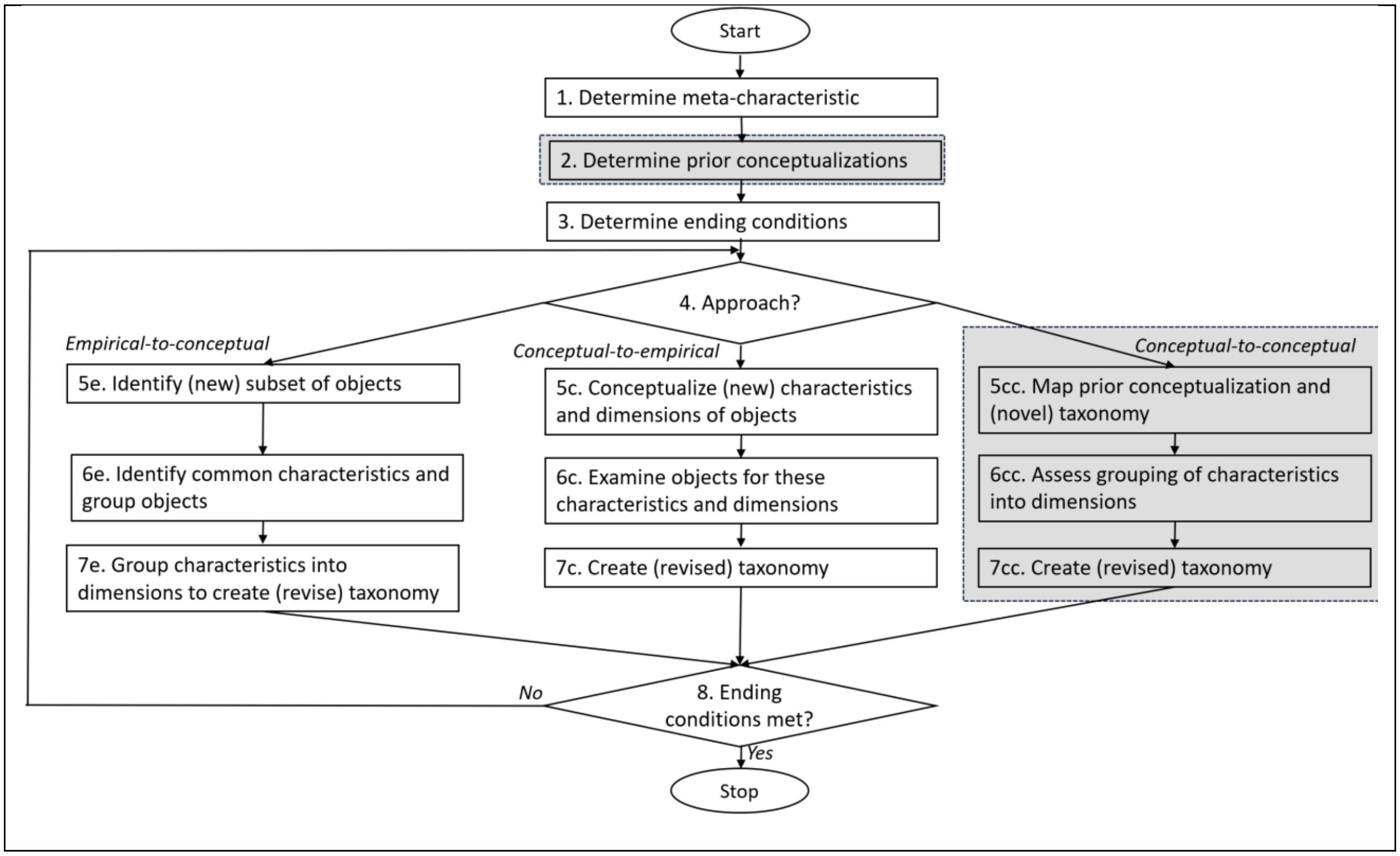

Figure 1. Extension of taxonomy development process of (Nickerson et al., 2013). The novel conceptual to conceptual steps are shaded.

After the c2c iteration, we conduct an e2c step in which we explicitly revisit literature, searching for empirical data for those concepts we identified in the c2c iteration, specifically we expanded on the people dimension, but also on aspects such as data quality, diversity and aligning AI to humans. This search was conducted in a narrative manner (King & He, 2005), where we searched with variation of concept names filtering either by year (after 2022) or by restricting towards GenAI, e.g., “Generative AI AND data quality”. The reasoning being that a systematic review on each of the concepts would be barely feasible within a manuscript and would add limited value as for each concept stringent selection would have to take place on what works are included.  Second, a narrative review allows to enhance the concepts while acknowledging the rapid pace of development. In turn, we performed another e2c iteration, which led to the introduction of “corporate value alignment” and a distinction between individual usage skills and GenAI as part of a team. In a revision of the framework another c2e and e2c cycle was performed with works from 2025. It led to the introduction of “scope monitoring” and “adaptive governance” as part of procedural mechanisms accounting for the fast paced change of technology and regulation. Also, the (narrative) search yielded many technical works and led to the addition of technical mechanisms to reach targets and enable governance. Another c2c iteration also yielded that prior frameworks did not account for intricate cost-benefit trade-offs that are essential to govern for organizations, which led to specific searches related to such trade-offs (e2c iteration). In this final iteration, we complemented the Google Scholar search with ChatGPT 5.6 Pro (Extra High) to identify potentially relevant literature (Wagner et al., 2026), which was subsequently scanned by the authors. Another c2e and e2c cycle in 2026 did not yield any changes to the framework, though a significant number of references were introduced and updated (easily visible through all 2026 additions in the references). After this no more changes were made and ending conditions were met.

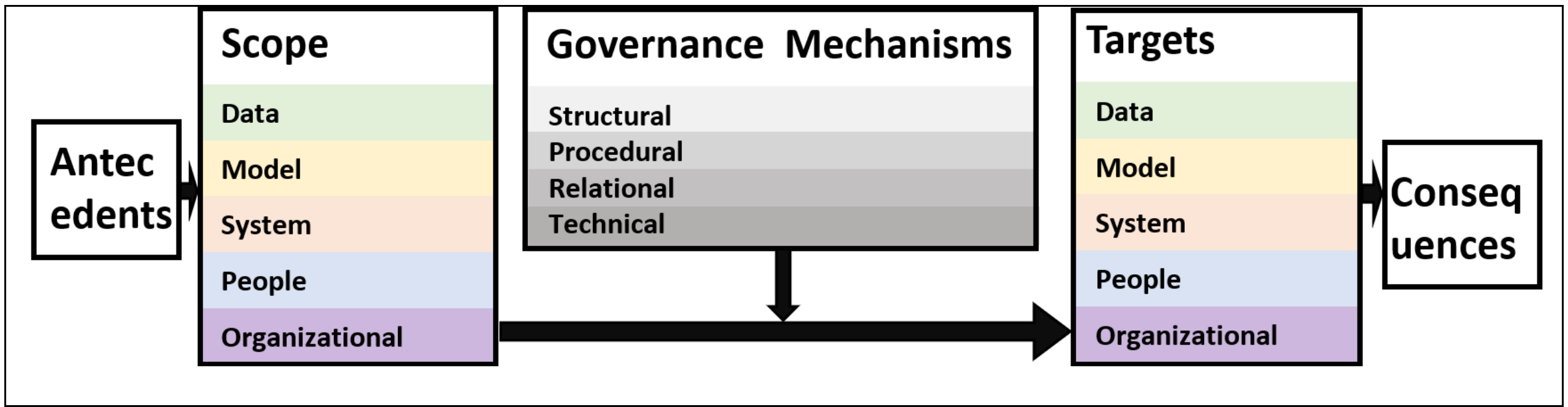

Figure 2. Framework dimensions

## Results

To contribute to gap-filling conceptualization(Leidner, 2018) and following adjusted taxonomy development process of (Nickerson et al., 2013), we structure the findings in the framework illustrated in Figure 2. These are antecedents (pre-existing conditions), scope (objects of governance) and targets (measurable properties related to data, model, system, people, and organization), governance mechanisms (structural, procedural, and relational levers to impact targets), and consequences (lagging outcomes). In the end, Figure 4 shows the complete framework. For each dimension, it specifies characteristics that require consideration. In contrast to prior work (Schneider et al., 2022), we include *people* as a key scope because bottom-up initiatives, user prompting and human-AI collaboration (also pointed out as risk in NIST, 2024) are central aspects of GenAI. Focusing on people has also been a key aspect of practitioners' and academics' works (Scott Likens and Nicole Wakefield, 2023), (Sigfrids et al., 2023) as well as the EU AI Act (EU, 2024). In addition, our work is also conceptually simpler than Schneider et al. (2022) as it directly links scope to targets through governance mechanisms. Aside from people as a new dimension, each existing dimension (data, model, system and organization) is expanded with multiple novel characteristics, e.g., for scope we added modalities, purpose, customization, failures etc. In addition, also existing characteristics were refined, e.g., for model training we enhanced supervised and unsupervised learning with self-supervised and reinforcement learning.

### *Antecedents*

Antecedents are conditions influencing the adoption of governance practices. We present them grouped by internal conditions and external conditions.

Internal conditions: We identify five relevant internal conditions, namely organizational culture, AI capability and costs, task structure, organizational strategy and change capacity. If a *culture* lacks openness to try out new tools based on GenAI (Dencik et al., 2023) or sufficient *AI capability* (in particular of staff (Hughes et al, 2026)), experimentation and skills to leverage employees' potential to innovate might be supported by targeted governance mechanisms such as training and incentivization. *Strategy* and existing governance efforts influence the objectives and constraints of GenAI governance efforts. For example, if an organizational strategy emphasizes customer trust, GenAI might not be rolled out among customer groups who are known to be more distrustful of AI. While GenAI can potentially be valuable for any task, (McAfee et al., 2023) point out that *task structure* is an important factor in capitalizing on GenAI. They propose that for knowledge-work jobs two key questions should be asked: (i) "How much would an employee benefit from having a competent but naive assistant who excels at programming, writing, or summarizing information but knows nothing about the company?" and (ii) "How much would an employee benefit from having an experienced assistant who has company specific knowledge?" The answer to these questions impacts implementation choices such as whether GenAI is used in a customized or general version, or whether it is implemented on-premise or using a cloud service. Respective choices impact aspects of risks and governance. For instance, IP leakage through user-AI interaction is less of a concern for on-premise systems accessible only within a company than for cloud systems hosted by 3rd parties with public access. Fifth, organizational *change capacity* (Hughes et al, 2026) could mean restructuring workflows or systems. Change issues circle around resistance to cultural change at the managerial level along with challenges in aligning AI systems with existing infrastructure.

External conditions: We identify three relevant external conditions, namely country, law and regulation, and industry. *Country* and language in which an organization operates impact how organizations effectively tailor and localize GenAI applications, with an awareness for potential performance differences across different markets or user groups. While all leading models support a variety of languages, research shows that GPT-5 for example performs worse on standard tests, if the test is translated and administered in different languages (OpenAI, 2025). This divergence can be explained with differences in the amount of training data available in different languages. Moreover, research indicates that the model security can be affected by the language used, as usage of low-resource languages is somewhat more likely to intentionally or unintentionally circumvent model safety measures, causing jailbreaking challenges(Deng et al., 2023). The relevance of training data availability extends beyond language to other country-specific conditions. For example, small countries like Liechtenstein with less than 50000 inhabitants and a low crime rate have only a few court rulings that can be accessed by a GenAI, therefore making the development of country-specific legal applications more difficult. *Legal and regulatory requirements* restrict AI usage. As of now, the regulatory framework addressing GenAI is still at a formative stage, resulting in significant unpredictability for organizations. There is an ongoing debate on how to regulate systems, how to distribute accountabilities, and what data can be used to build GenAI. For example, ChatGPT was temporarily banned by regulators in Italy (McCallum, 2023). Research finds significant divergence in compliance with the requirements of the EU AI Act across different GenAI model providers, while pointing to a lack of standardization for compliance evaluation (Guldimann et al, 2024). Thus, adopting any of them potentially constitutes a compliance risk for organizations. Divergence also occurs with respect to regulations globally, e.g., the EU adopts a stricter, proactive regulatory approach, while the US is more reactive and lenient (Symeonidis et al., 2026).

One ongoing regulatory debate related to GenAI concerns the issue of copyright (Grynbaum & Mac, 2023) which holds implications for organizational risks of copyright infringements, and the management and enforcement of IP rights. In the context of training GenAI, relevant questions include whether certain data can be used to train the models, whether outputs can be used, and, if so, if remuneration applies. Common data collection techniques, such as crawling from internet sources might lead to the inclusion of copyrighted and private data (Grynbaum & Mac, 2023). With respect to GenAI deployment, regulatory uncertainty remains regarding the protectability of model outputs and prompting instructions, which affects how organizations can manage and enforce related IP rights. To present an example, (Murray, 2023) explored the connection between generative art and non-fungible tokens (NFTs), pointing out that under existing U.S. copyright regulations, wholly generative art is not eligible for copyright protection due to its creation by non-human entities, but significant human involvement might allow for copyright (U.S. Copyright Office, 2025), though copyright laws vary globally (Furqan et al., 2026). Besides (inter-)national regulatory frameworks, the usage policies of GenAI providers influence the governance within organizations. For example, by default and to support certain features, users of the free version of ChatGPT provide their inputs and outputs to OpenAI for further model improvements (OpenAI, 2024). In this way, proprietary or sensitive personal data could be effectively included in and retrieved from (future) systems. The policies of model providers therefore relevantly impact risks related to data rights, data protection, and IP leakage. GenAI governance within a specific organization is furthermore influenced by *industry* conditions. First, industry-specific regulatory requirements might exist. For example, applications in the healthcare sector are likely to see stricter regulation, such as requirements of human oversight, compared to applications in creative industries due to the gravity of risks of potential medical errors (EU, 2024). Second, the competitive dynamics in an industry can dictate the pace of GenAI adoption, with potential implications for governance strategies. Third, varying levels of technical maturity across industries can impact the ease of integration and business value of GenAI, demanding different levels of investment and expertise which are likely to reflect in governance mechanisms.

### *Data scope & targets*

Data scope: With respect to data, GenAI governance includes considerations of data type, source, modalities, label, purpose, distribution, and the environment. *Data types* can be categorized into structured and unstructured, with GenAI mostly being trained on unstructured data such as text, images, or audio. Video generation and understanding are also increasingly emerging (Liu et al., 2024; Reid et al., 2024). GenAI can also handle structured data (Sui et al., 2024). Commercial vendors integrate GenAI in tools for analyzing structured data, e.g., Microsoft Excel and Power BI with LLM integration (Microsoft, 2024). Governance must consider the type of data involved in each GenAI use case because it dictates data management practices including solutions to ensure data integrity, compliance, and security. Turning towards *data sources*, sources can be distinguished based on whether the creator is human or not, i.e., an AI model or a machine, and whether the creator is internal (within the company) and external (third-party providers). Also, commonly data, in particular AI-generated content (AIGC) might emerge from a set of different interacting creators and sources, e.g., an AI model might generate an artifact based on a human prompt, which includes uploaded documents from different sources (Li et al., 2025). From a legal perspective the distinction between human and non-human data is relevant. Especially, personal data is more strongly regulated under instruments such as the GDPR. The importance of large amounts of (general) data for training GenAI drives the proliferation of data marketplaces and the importance of third-party data providers as seen with ChatGPT-5 (OpenAI,2025). Assessing and enforcing properties of data by third-party providers, such as data quality and timeliness, requires different mechanisms and can be more challenging compared to data generated within a company. For example, due to the lack of control and transparency of third-party providers (NIST, 2023) (Sec. 1.2.1). In the context of GenAI, the aspect of data sources also links to legal concerns of copyright infringements as generative models might replicate protected works that were included in the training data (Carlini et al., 2022). With respect to *modalities*, many state-of-the-art models such as GPT-5 are multi-modal, i.e., they are trained on and can process and generate different forms of data including text and images. Multimodality poses governance challenges including the definition of effective benchmarks to capture performance across different modalities (Weidinger et al., 2023). However, respective challenges can also arise in unimodal models, such as language-only models, which generalize to different tasks and input, such as dialogue, narrative and expository texts, or source code. The type of data used can impact model capabilities, e.g., training language models on source code is attributed to better reasoning skills (Madaan et al., 2022). In the context of GenAI, the *purpose* of data is manifold affecting its management. Instructions in the form of prompts can guide a model to perform company specific tasks. Data might also serve the purpose to improve various model metrics through training or retrieval and amendment to prompts (RAG). In traditional AI, training data is typically selected to ensure high model performance. For GenAI, there exist multiple training data sets with different goals aside from task performance, e.g., assuring ethical behavior and human-AI alignment (Ouyang et al., 2022). Commonly, ethical alignment is improved through fine-tuning an existing model through human feedback, e.g., reinforcement learning from human feedback (RLHF). For example, humans can rate the output of a language model as harmful or helpful and the model can learn from such feedback to optimize future outputs (Ouyang et al., 2022). In RAG, data also supports model grounding, auditability, and explainability(XAI) (Schneider, 2024). Data can also be split into *labeled and unlabeled data*. For GenAI labeled data is often used to fine-tune foundation models for specific tasks and for human-AI alignment. With GenAI, another type of data, previously known mainly from reinforcement learning, has increased in importance (Schaal, 1996): Demonstrations or "solution guides", which can be helpful for training (and for prompting). The chain-of-thought paradigm has yielded significant performance gains on standard benchmarks: A model is asked to "Think step-by-step" (Kojima et al., 2022) or is given input-output samples with a demonstration on how to derive the output from the input. Data exhibits a certain *distribution* that determines its suitability as a source for learning to solve a task. Relevant properties of the data distribution are, for example, balance (e.g., representing entities of interest by a similar number of samples), and completeness (e.g., all entities of interest are present). Completeness can also refer to single samples, e.g., the description of an entity containing all relevant attributes. Finally, as language, and data (distribution) in general, can evolve, being in a static or dynamic *environment* strongly impacts governance, e.g., the need to monitor and update GenAI.

**Data targets**: With respect to governance targets, *data quantity and quality* are important goals. Governance targets related to data quality include timeliness, correctness, and little noise (e.g., an entity is accurately described by non-outdated data). The difficulty to manage these targets is illustrated by OpenAI's ChatGPT, which – depending on the version – lacks data from the last couple of months up to years. The importance of data quality is also emphasized in the context of fine-tuning as research shows that high quality can reduce the number of samples needed to optimize LLMs for a certain task (Meta, 2024; Zhou et al., 2024). Additional data targets in the context of GenAI include aspects like avoidance of near duplicates to improve model robustness and reduce memorization (Lee et al., 2021), filtering toxic and harmful data, and assuring an adequate alignment of training data distribution and "test" data distribution. That is, the data used during training must adequately resemble the data used while operating the system, to avoid *biases* and overfitting, and assure model performance. For example, if some entities (such as specific ethnicities) are underrepresented (non-balanced data) the model might be biased. The *value of data* is generally difficult to assess and for GenAI a few technical attempts have been made to measure the impact of data on outputs (Yang et al., 2023). Data value is tied to data properties such as representativeness, diversity, and accuracy, as well as economic properties such as cost of collection and the revenue generated by model-driven services. For example, diversity can increase value, as it has been observed that an LLM fine-tuned only on a specific phrasing and type of question, e.g. "How to", performs worse than if tuned on more diverse questions (Zhou et al., 2024). Feedback from users, for example on output correctness, harmfulness, or toxicity, is often considered valuable and important to collect. Increasingly GenAI itself performs data quality management (LLM-as-a-judge), which should be supported by dedicated human oversight roles, e.g., related to data integration and resolving data issues (Roht et al., 2026). *Data security* constitutes another important governance target. Aside from standard security mechanisms from the pre-AI area, data security covers aspects that are closely interconnected with the model, e.g., the risk of data breaches when training data can be extracted by attackers during model deployment (Nasr et al., 2023). That is, GenAI models might allow access to sensitive data used for training using ordinary prompting without access to model internals. Governance also includes targets related to the *data lifecycle* which have been discussed with a focus on privacy and copyright (Zhang et al., 2023). This includes managing consent modifications and withdrawal, which may necessitate retraining models. Relatedly, and linked to the issue of data timeliness, *data drift* mitigation is a critical concern. Data drift refers to the gradual change of input data over time, leading to a gradual outdatedness of the training data known to a model. As many GenAI use cases involve recent information, data drift can happen quickly. At the same time, preventive strategies are costly given the price tag of continuously integrating large amounts of new data through model re-training with adequate, high quality training data. Lastly, as data sources become larger and more diverse and regulation advances, complete and accurate *metadata* including *data lineage* become important governance targets. Metadata provides information about data provenance, usage, and parameters. It is thus important for documenting legal compliance and ensuring accountability, e.g., to identify which data source owner and creator are responsible. With respect to data lineage, i.e., the origin and processing history of data, important aspects include user consent and control over their data (Newman, 2023). First attempts have been made to provide tools for data provenance and curate licenses of open-source datasets (Longpre et al., 2024).

### *Model scope & targets*

Model scope: With respect to models, GenAI governance includes considerations of training, customization, learning, hosting and development, model failures, alignment, and vulnerabilities. Fine-tuning resembles one form of *customization* in which a pre-trained model, i.e., its parameters, is adjusted towards a domain specific task. Alternatively, customizing GenAI to a specific task can be achieved through prompt engineering, i.e., model inputs that are crafted to optimize task-specific model behavior without changing the model itself. Technical customization benefits from organizational capabilities, such as strategic data management and a flexible GenAI stack (Diaferia et al., 2026).

Prompting can be done manually or also in an automatic manner using so called “soft-prompting”, where a training process leads to virtual tokens prepended to a model’s user inputs, which tends to lead to prompts outperforming manually designed ones for a specific task (Liu et al., 2023). Prior to GenAI, raw models were adjusted, i.e., trained using supervised learning. GenAI has seen the rise of self-supervised learning, which is a form of unsupervised learning, as it does not require labeled datasets, but labels are generated based on auxiliary tasks. For example, for training LLMs, the next word in a text commonly serves as a label. Furthermore, reinforcement learning has been used to optimize model behavior (Ouyang et al., 2022). Generally, the training process of models has become more intricate involving potentially self-supervised learning on very large amounts of data, followed by instruction fine-tuning, human-AI alignment, and supervised task-specific fine-tuning on smaller data. One classically can distinguish between online *learning* and offline learning, which refers to whether model updates occur continuously from observed data or from time to time. This boundary has blurred as GenAI can maintain an awareness for context, such as the preceding conversation, and leverage additional information, e.g., retrieved from the Internet. Especially, GenAI saw a novel paradigm called “in-context” learning that does not alter the model, but rather the model benefits from in-/output samples that help the model understand how the output should be (Brown et al., 2020). The *development and hosting* of models differ significantly for GenAI. Compared to previous machine learning systems, GenAI models are significantly larger, complex deep learning models that are typically based on foundation models (Bommasani et al., 2021), i.e., mostly transformers (Banh & Strobel, 2023; Vaswani et al., 2017). Training these large-scale models is extremely costly due to computational demands and the need for vast amounts of data. Consequently, development and hosting of foundation models is centralized with a few providers. When deploying GenAI, most companies will therefore rely on externally developed pre-trained models, possibly adjusting them to their specific needs rather than internally developing models from scratch (Schneider et al., 2024b). Hosting and customization can be done internally or externally, e.g., in the cloud. Some (customized) models like OpenAI’s ChatGPT can only be hosted and customized in the cloud as of now. The large scale and opaqueness of foundation models critically relate to the aspect of model *failures*. Critical failures include reasoning errors and the generation of fabricated, untrue facts, referred to as hallucinations (Kalai et al., 2025). Hallucinations are particularly difficult to spot as GenAI, such as LLMs, tend to communicate fluently but non-factual. *Alignment* means that model outputs align with the user, organization and legal intent or, more formally, the degree of overlap of different outcomes of two agents (e.g., the model and the user) (Askell et al., 2021). While alignment has been defined focusing on humans as agents, it can be generalized, i.e., in a corporate setting, agents include any stakeholders and the organizations as a whole.

The prevalent scope includes helpfulness (do what is the agent's best interests), honesty (convey accurate information and avoid deception), and harmlessness (avoid harming agents) (Askell et al., 2021). For example, in (Schneider et al., 2024a) architects criticized AI's failure to follow precise requirements and its tendency to mirror issues (already identified and mentioned by users) when being asked for feedback. Organizational intent issues might refer to, e.g., failing to prevent erotic content suitable in private settings. Multiple GenAI applications show emergent properties that arise from an increase in model size, data, and compute without intentional model training (Brown et al., 2020). As these capabilities cannot be anticipated during development, their potential harm and implications for model performance in a specific context is difficult to assess a priori (NIST, 2023)(Sec. 1.2.1). GenAI has suffered from (security) *vulnerabilities* that have been exploited. Knowledge has been extracted from commercial models to fine-tune other models: In the case of the Alpaca model querying ChatGPT for less than 600$ was sufficient to extract knowledge to train the model to perform better than the original on some human evaluations (Wodecki, 2024). User instructions and data might conflict with system instructions, potentially causing models to ignore safety and moderation instructions also known as jailbreaking (Deng et al., 2023). Potential misuse scenarios for GenAI include social engineering attacks (psychological manipulation of people to nudge them into performing actions desired by the attacker), phishing attacks (posing as trustworthy entities to obtain sensitive information), support hacking and ransomware (e.g., through automated code generation) (Gupta et al., 2023).

**Model targets**: With respect to governance targets, model lifecycle, quality, alignment, safety and security are important. Beginning with *Model lifecycle*, it covers aspects such as model monitoring and (fast) updates (due to user feedback or uncovered ethical or security concerns) and efficient and comprehensive evaluation. Related to *quality,* hallucination and reasoning shortcomings that compromise output correctness are innate problems of LLMs, as a subcategory of GenAI, that to this date can only be partially mitigated, e.g., by using retrieval augmented generation (RAG) (Lewis et al., 2020). RAG retrieves information with relevant facts for a prompt and constrains models to focus on the retrieved information. It reduces the risk of fabrication of facts and facilitates human fact-checking of model outputs. Output correctness is further challenged by the limited robustness of GenAI applications that arises from their sensitivity to user prompts. That is, small changes in a prompt can lead to large changes in the model output (Lu et al., 2021). Performance control of GenAI is also challenged by its generative nature. As GenAI produces artifacts rather than decisions, novel metrics are needed to assess model performance, including metrics focused on individual outputs rather than the model. For example, if a GenAI generates a story or a piece of code, metrics for these instances are more interesting than for the model in general (Sun et al., 2022). Quantitative evaluation of models is challenging, as common metrics such as BLEU and ROUGE score for LLMs are useful but inaccurate (Celikyilmaz et al., 2020). Related to the controllability of correct outputs, model explainability (XAI) constitutes another important target of GenAI governance. Different approaches to GenAI explainability exist. For example, models can be requested to produce and report outputs in a "step-by-step" manner, which can serve as human-comprehensible explanations though they are not guaranteed to be faithful to actual model behavior. However, despite tremendous effort to address XAI, it is still not solved for AI and even less so for GenAI (Longo et al., 2024; Schneider, 2024). For GenAI in the context of code generation (Sun et al., 2022) a number of XAI needs arose from a user study, i.e., system requirements and impact to anticipate suitable conditions for use and limitations of the model's capabilities. The paper also discusses adaptations of prior and novel XAI features. For example, users were interested in understanding alternate outputs and saw explanations as a process that could be either human or AI-initiated. Also, users were interested in obtaining information on model prediction uncertainty and requested features on how to resolve such uncertainty. Another governance target related to quality is model efficiency, i.e., how much computation is needed to obtain output. Better efficiency contributes to common sustainability goals, e.g., lower energy consumption, but also implies lower computational demands, i.e., lower costs and potentially faster response times. There is often an efficiency-accuracy trade-off (Xu et al., 2023). A second governance target is *AI alignment*, that is ensuring a congruent behavior of a model with human and organizational interests, and ethical and legal expectations. Targets in this context can be use-case specific. For example, in the healthcare domain empathy might constitute a desirable model property, referring to the ability to understand the personal experiences and emotions of another, without extensive bonding (Oniani et al., 2023). One study compared ChatGPT's responses and those of physicians, showing that ChatGPT was preferred by healthcare evaluators in ¾ of all cases in terms of quality and empathy (Ayers et al., 2023). With respect to ethical and legal alignment, GenAI governance must address concerns such as model misuse, generation of harmful or inappropriate content, widening of the digital divide, and the reproduction of bias with respect to gender, cultures, and minorities (Fui-Hoon Nah et al., 2023).

Avoiding these issues is a high-stakes alignment goal. Regulatory compliance requires, for example, extensive documentation on respective preventative measures for high-risk GenAI applications (Hupont et al., 2023). A third governance target is *safety, security and privacy* which relates to the aspect of ethical and legal alignment. Models with access to company-specific data must be secured like raw data. GenAI governance must assure that models do not cause danger or harm. This includes preventing the dissemination of harmful, illegal, and unethical content and protecting against leakage of personal or proprietary data. Moreover, governance must address the risk of model misuse.

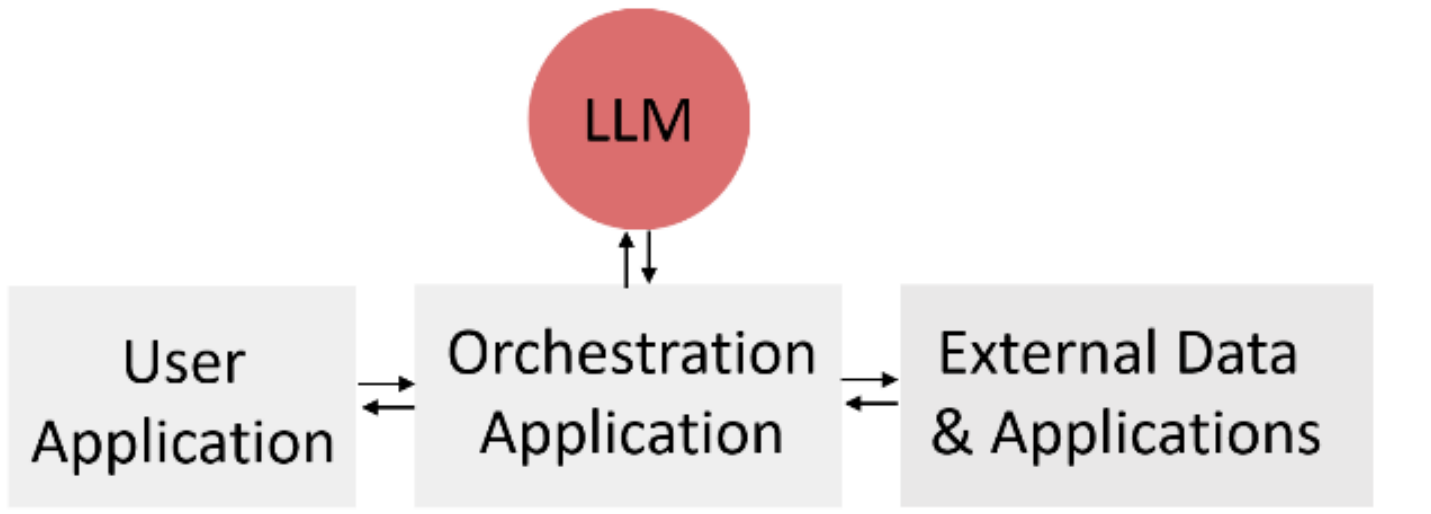

Figure 3: System architecture with LLMs

***System scope & targets***

System scope: With respect to systems, GenAI governance must address a complex architecture involving numerous interacting applications and data sources, the degree of system autonomy and human interaction, and the model scope. Figure 3 illustrates a common system *architecture* centered around an orchestration application. In response to prompting instructions, the orchestrator might obtain data from an external data source, e.g., from a database or the internet, to provide the LLM with prompt-relevant information (Lewis et al., 2020). GenAI systems may also leverage other tools, e.g., translate a natural language question into code, run the code using a conventional code interpreter, and package the result of the code interpreter in a human-understandable text presented to a human (McAfee et al., 2023). GenAI systems can serve as a facilitator for *human interaction*, e.g., translating user inputs in natural language into formatted API calls as demonstrated by OpenAI's Assistant API (OpenAI, 2023d). However, GenAI systems also have the potential due to their unprecedented often super-human performance (as already witnessed by ChatGPT-4 (OpenAI, 2023a)) to act more *autonomously* and to conduct novel tasks that were not possible previously. The degree of autonomy impacts governance needs, e.g., more autonomous systems need more careful monitoring, e.g., in the form of guardrails (Dong et al., 2024), which monitor and filter the outputs (and inputs) of models. For example, the open-source toolkit Nemo (Rebedea et al., 2023) allows to specify topics such as politics that an LLM should refuse to discuss. Furthermore, core functionality of systems stems from models, the *model scope* also partially extends to systems. For example, a company might host a model at a cloud provider, but the system leveraging the model is hosted on-premise or vice versa. Integrating GenAI into enterprise platforms may require coordinated architectural and governance adjustments to maintain stability (Safaei et al., 2026).

**System targets**: Targets include performance, facilitation of human-AI collaboration, and accountability and transparency and model targets. The governance targets addressing the system scope are entangled and overlapping with model targets.  For example, model security can be partially ensured with the help of system security, e.g., login mechanisms can help to link attacks to users and, in turn, act on them, and guardrails. System security connects to the aspect of *system performance* and includes concerns like resilience. Resilience refers to a system's ability to withstand unexpected adverse events. Mechanisms to disable AI while ensuring that a system remains operational as a means of risk mitigation, were discussed, e.g., for healthcare (Nicholson Price II, 2022). Elements such as RAG that access external knowledge also pose a risk, e.g., if this data is manipulated (Zou et al. 2025). With respect to *human-AI collaborations*, targets include making a system accessible to a large group of people rather than just experts such as system developers or computer scientists. This can be achieved for example through accessible interfaces and low- and no-code tools for system usage (Solaiman, 2023). Lastly, two governance targets are *accountability* and *transparency* of a system. Accountability (Oniani et al., 2023) requires tracing activities on a system to individuals who may then be held accountable for their actions (Oldehoeft, 1992). In this context, accountability means that actions of a GenAI can be attributed to specific entities, including organizations, developers or individual users, who can be held responsible (Horneber & Laumer, 2023; NATO, 2021). Accountability requires a degree of traceability, including tracking and documenting the data, processes, and artifacts related to a system (Mora-Cantallops & others, 2021).

Traceability also serves the governance target of transparency, which is important for example for fostering user trust and compliance reasons.

### *People scope & targets*

People scope: With respect to people, the GenAI governance scope encompasses *stakeholders* engaged in planning, developing, and using the system. This includes data, machine learning and software engineers, project managers, people observing the system after deployment, and end-users. Moreover, the scope of people also covers groups, communities, and societies that can be affected by the transformative impact and risks of GenAI (NIST, 2023). People will, in the role of end-users, be progressively exposed to GenAI when using ordinary software applications. The integration of Microsoft's Copilot built on OpenAI's technology in Microsoft 365 spreadsheet, word processing and presentation software exemplifies this. Through such progressive integration and the likely development of systems increasingly capable to perform complex tasks autonomously, GenAI can evolve towards the role of a virtual team member. GenAI can boost performance of individuals to reach that of teams without AI, it can help bridge functional silos and it can increase positive emotions (Dell'Acqua et al., 2026). Stakeholders might perceive changes due to GenAI positively or negatively and, more generally, differ in their *expectations* towards GenAI. GenAI has caused work alienation, e.g., for marketing and customer service tasks in tourism (Hai et al., 2025). More broadly, more than 1/3 of more than 100 surveyed CEOs say that AI could destroy humanity within a decade (Matt Gean, 2023), while a number of scientists such as Geoffrey Hinton also warn of the dangers of AI, instilling fear among people. Trust in AI and its perception of usefulness are predictors of intentions to use GenAI at work, while factors such as gender and education are not (Nguyen et al, 2026).

*Processes* and the *role* of humans might change from working alone (or with fellow employees) towards collaborating with AI (Microsoft, 2023). Human-AI teaming (Vats et al., 2024) is likely to grow in relevance and already optimizing human-AI collaboration is seen as a key factor to maximize AI's effectiveness(Calvino et al., 2025). AI can take very different roles as teammate, e.g., AI can be "a second pair of eyes" (Henry et al., 2022) or provide stimuli in joint brainstorming (Memmert & Tavanapour, 2023). Within a company every employee might become a supervisor of an AI assistant. The employee must carefully assess the AI's outputs as illustrated in the case of building design, where the AI might create designs that must be checked for feasibility by architects (Schneider et al., 2024a). The line between different stakeholder groups is blurring in the context of GenAI, as developers, end-users and system adopters can significantly impact GenAI behavior through prompt engineering and easy to add knowledge to GenAI systems using RAG as demonstrated by OpenAI's custom GPTs (OpenAI, 2023c), which allow users to build and share customized ChatGPT versions with a few clicks. Roles might become broader, where GenAI enables tasks assigned to external providers (Schmidt & Schöbel, 2026). The goal of prompt engineering can be improvement of model performance on a specific task, leveraging in-context learning, or customizing a model for a specific audience. For example, by instructing a model to "Translate every input to German. If it is in German translate it to English." one has effectively built a translation service where any user input will be translated. While interaction with GenAI is simple, prompt engineering, especially for safety-critical system prompts is non-trivial and a skill requiring mastery (Khurana et al., 2024; Zamfirescu-Pereira et al., 2023). Besides their exposure to, and impact on GenAI, stakeholders must also be considered in their respective relevance for security and safety risks. Users, including company employees, might use GenAI unethically, for example by requesting the generation of harmful or toxic content. While measures to prevent unethical outputs are integrated in leading models, undesirable results may still occur. This can happen due to failures in classification and filtering, or because the model's ethical standards differ from those of the company. In this context, especially given the sociocultural subjectivity of ethics, there is also debate over who is responsible for – that is, who may take charge of – determining the ethical standards implemented in GenAI. For example, while commercial systems such as OpenAI's ChatGPT have relatively strict ethical guidelines, open-source models may have fewer constraints making it easier to obtain harmful and unethical content as demonstrated by GPT4chan, which puts the responsibility to uncover and restrict unsuitable responses on end-users and deploying companies. However, individuals might struggle to evaluate the ethical and factual correctness of AI outputs; a challenge that is intensified by the tendency of models like ChatGPT to fit their response to (incorrect) user assumptions instead of clarifying them (Tan et al., 2023). Intellectual property (IP) and privacy risks are another reason to include people in the scope of GenAI governance as employees can leak confidential data when using third-party GenAI. For example, employees at Samsung entered confidential information into ChatGPT while using the AI for code debugging. OpenAI can integrate such data into the training data, potentially making it accessible for future public responses. This illustrates the significant privacy and IP concerns governance must address in the context of GenAI, and the respective role of individual employees. For instance, if (input) data is gathered and integrated in (future) models, companies require a policy prohibiting employee use of respective model, as this can lead to information leaks and data protection breaches (Gillespie et al., 2025).

**People targets**: Governance targets encompass people's attitude towards GenAI, skills to use and assess GenAI, and an awareness of its capacity and limitations. A company might aim for people trusting AI and perceiving it as useful (Nguyen et al., 2026) to support adoption, while also having a critical stance in order to reflect on AI's outputs (rather than blindly trust them) and in order to engage in experimentation potentially leading to bottom-up innovation. Li et al. (2026) identify mindfulness and job crafting as human-centered conditions associated with effective GenAI adoption in project work. Beyond shaping people's attitude, technical and practical ethics training are critical and should be implemented not only within organizations, but already at early-stage education (Solaiman, 2023). Besides ethical awareness, users must be trained in critical thinking to reduce the risks of false trust and overreliance on GenAI outputs (Lee et al., 2025). Employees require skills to assess GenAI outputs for their ethical alignment with corporate values and correctness. Moreover, they must be aware of possible IP risks associated with GenAI usage. Besides risk reduction, facilitating employees' effective use, i.e., generating correct and relevant outputs, is another key target for leveraging the business value of GenAI (Khurana et al., 2024; McAfee et al., 2023). While interacting with LLMs is simple, there are several strategies to improve model outcomes(von Brackel-Schmidt et al., 2023; White et al., 2023). For example, (Liu et al., 2023; Schulhoff et al., 2024; White et al., 2023) proposed catalogs of prompt patterns. Companies are advised to educate on respective knowledge as not all employees can be assumed to handle prompts well (Zamfirescu-Pereira et al., 2023). An overview of GenAI competencies is provided by (Annapureddy et al., 2024).

### *Organizational scope & targets*

Organizational scope: We distinguish between an intra- and inter-organizational governance scope. Governance within an organization (*intra-organizational*) is more challenging compared to the pre-GenAI era. GenAI has broader applications and more complex risks. It therefore necessitates more extensive governance efforts including technical, domain, business, and legal experts. Governance must extend beyond technical artifacts (such as data, systems, and models), for example, individual employees who can easily access freely available GenAI, which might be tempting (even if forbidden) but comes with risks as elaborated above. Moreover, due to the multipurpose nature of GenAI, governance of model usage can no longer be task-specific but rather requires a consciousness for content, timing, and context. For example, while using ChatGPT to improve a product marketing text might be acceptable, employing it for a patent application might not, because it poses a higher risk of IP leakage. Furthermore, processes might change in profound ways, e.g,. GenAI can become a structural collaborator shifting sub-tasks towards one person with fewer handoffs as shown for startups (Schmidt & Schöbel, 2026). For instance, AI can empower clients to take over parts of the value creation using GenAI (Schneider et al., 2024a), posing additional challenges for companies such as revenue loss and being confronted with AI's shortcomings such as hallucinations that might not be recognized by clients and employees. GenAI can shift decision influence from human experts toward AI-generated outputs, as these outputs increasingly shape which information is considered credible and which knowledge claims are prioritized (Alshammari, 2026). However, formal decision rights and oversight might still demand human experts, leading to a governance tension.

In the context of GenAI, *inter-organizational* governance includes additional actors that are integral to the new AI supply chain (Schneider et al., 2024b). These include foundation model providers and actors offering downstream applications built on third-party foundation models for deployment. An increasing number of actors contributes to the development of a specific GenAI application. Thus, control and responsibility to mitigate model-inherent risks diffuse across this inter-organizational network requiring potentially triadic arrangements (Plugge et al., 2026). As elaborated above, concerns include ethics, security, and safety risks, as well as performance concerns. While even the providers of foundation models struggle to strictly prevent undesired system behavior, their governance in form of usage policies limits the governance options of deploying companies. Hosting models externally is nevertheless appealing to avoid technical challenges and to lower investments.

*Costs & benefits* of GenAI are often intertwined between data, model, systems, people and organizational aspects leading to a wide range of trade-offs illustrated through a set of examples. Costs can be direct or indirect. Direct costs cover for example costs for computation such as tokens/$ (Brodsky, 2024), data storage, e.g., GB/$, AI development, while indirect costs can arise from reputational damage due to AI failures, changes in employee work satisfaction due to GenAI(Hai et al., 2025), rework or slow-down due to low quality AI response(Becker et al., 2025), etc. Benefits are often domain specific and include increased productivity, overall cost reduction (e.g., due to smaller teams and AI replacing external experts in startups (Schmidt & Schöbel, 2026)), faster services, improved quality (McAfee et al., 2023; S. Noy & Zhang, 2023). However, only a minority of organizations report EBIT effects (McKinsey, 2025). A fundamental trade-off is model costs vs. response quality and duration. Larger models require more computation (and thus lead to higher costs) but also lead to better outcomes (e.g., in terms of prompt following and reduced hallucinations) as illustrated through benchmark results (Meta AI, 2025). (These) benchmark results also show that newer models often perform better with the same amount of computation, e.g., due to better training procedures, but replacing existing models integrated into systems might cause significant costs due to adjustments of prompts and human evaluation. In the case of reasoning models, which plan and reflect leading to a large number of self-generated model queries, the amount of computation and time to answer a query can significantly increase in addition to better outcomes. Poor outcomes can cause inefficiencies as outcomes need to be human-validated, altered and potentially additional prompts are needed. In extreme cases, GenAI might even reduce productivity (Becker et al., 2025). Furthermore, errors can lead to real-world costs, e.g., due to poor legal advice (Weiser, 2023). Increased computation time might cause idleness of employees waiting for AI's responses. Model performance can often be improved by model customization using task-specific data, which collection and cleaning might be costly. Implementing GenAI on-premise limits dependence on external providers, but can cause high costs for hardware and system development and maintenance among other trade-offs related to control, capability, and integration (Toorajipour et al., 2026). Similarly, human oversight might reduce legal risks but increase task duration and diminish benefits. A number of paradoxes have also been reported for GenAI chatbots for customer service (Ferraro et al., 2024). Chatbots can enhance customer satisfaction but also frustration. They are sufficiently powerful to revolutionize customer service but vulnerable to bad actors.

**Organizational targets**: With respect to intra-organizational governance, a relevant target is *governance alignment,* especially cross-functional alignment. As GenAI might emerge bottom-up from individual employees or departments, individual functions might develop their own governance mechanisms. While bottom-up governance solutions might effectively account for task-specific aspects, they should be aligned at the firm level and with firm-level concerns. Mitigating risks of GenAI usage within a company must be balanced against fostering a culture that invites experimentation and the identification of AI use cases to leverage the economic potential of GenAI (Dencik et al., 2023). This balancing act has been demonstrated for creative settings as "speed with fit", where governance enables rapid experimentation preserving accountability and authenticity (El Rassi, 2026). Employee-driven experimentation can generate valuable organizational knowledge (Sposato, 2026). With respect to inter-organizational governance, one target is harmonizing digital strategies including governance aspects within the ecosystem (Dencik et al., 2023; Scott Likens and Nicole Wakefield, 2023). In particular, where a company is not fully controlling a GenAI system internally but hosting it externally, governance alignment may only be achievable through contractual, not technical, measures (Reddy, 2024). Governance responses are also important to reduce (legal) risks of inter-organizational activities; however, regulation is currently evolving and does not yet clarify all aspects for inter-organizational usage. Companies generally strive for *cost efficiency* under multiple constraints and conflicting objectives often aiming for *pareto-optimal trade-offs* meaning that no better solution can be attained balancing both costs and benefits.

### *Governance Mechanisms*

Governance of GenAI encompasses structural, procedural, and relational governance mechanisms (Schneider et al., 2022). **Structural mechanisms** include organizational structures such as roles and the location of decision-making authority. Due to the high-risk potential of GenAI, it might necessitate specific *roles* such as a Risk Officer or Ethics Officer(Schäfer et al., 2022; Scriven et al., 2023), as well as committees encompassing domain, ethics, law and AI experts (Reddy, 2024). Experts should be involved when risk is high and should also include social scientists and multidisciplinary perspectives. In the context of distributed data governance for LLMs across 60 countries, roles such as data modelers, custodians and rights holders have been proposed (Jernite et al., 2022). Actors with expertise and leverage, including companies developing GenAI, must have an incentive to engage in governance discussions and be held accountable to commitments for safe releases. Leading GenAI labs publish safe release strategies. Google's position on responsible AI practices for example encourages in-house risk evaluation and mitigation. But conflicts of interest could result in dismissal (Ebell et al., 2021). The allocation of compute power has been promoted as another structural governance mechanism for companies developing GenAI (Sastry et al., 2024). When it comes to the (location of) *decision rights* as a means of governance, GenAI shifts relation from “the business depends on engineers to build AI” toward “engineers depend on knowledge owners, risk and compliance, and end users to steer AI” (Schmitt & Grauer, 2026). Inspiration can also be drawn from policymakers: different countries have taken different approaches to regulating AI. Vertical regulations target a specific application or manifestation of a technology. This contrasts with horizontal regulations, such as the EU’s AI Act, which are comprehensive umbrella laws attempting to cover all applications of a given technology. In China, a mix of both is visible (Sheehan, 2023).

**Procedural mechanisms** focus on the way decisions are made, and actions are taken. They include a high-level strategy that guides GenAI governance based on strategic business objectives. Prominent consulting firms suggest important elements of a GenAI *strategy* including management of AI risk, putting people centerstage, developing a productivity plan, and working with the ecosystem to unlock GenAI potential (Scott Likens and Nicole Wakefield, 2023). Generic pre-GenAI standards like ISO/IEC 42001 for AI management (ISO/IEC, 2023) provide generic procedures. Academic work has also stressed management strategies for GenAI adoption to yield strategic advantages, such as expanding existing competitive advantages with GenAI and identifying visionary business opportunities (Ruokonen & Ritala, 2025). The GenAI human resource management strategic framework by (Chowdhury et al., 2024) covers key steps from understanding to assessing, on to transforming. It also emphasizes a bottom-up approach viewing employees as intrapreneurs facilitating change. For AI risk management, a three-layered approach has been proposed focusing on capability (technical), human interaction, and systemic impact (society) (Weidinger et al., 2023). Other works discuss how to mitigate risks along the GenAI lifecycle, including prelease testing where users aim to identify flaws (Bell et al., 2023). Another strategic dimension relating to both AI risk mitigation and business strategy concerns the debate whether to release GenAI as open- vs. closed-source models. While only few companies train foundation models, it can be expected that increasingly companies fine-tune models and face the same decision. In this context, Solaiman (2023) distinguishes three types of release artifacts: (i) access to the model itself (ranging from fully open models, providing access to model weights only, allowing only to upload inputs and access outputs using an API), (ii) components that enable further risk analysis (e.g., evaluation results on benchmarks, information on training data, etc.) or (iii) model replication (details on training, access to data, etc.). Considerations of release include societal concerns (e.g., risk of exploitation of minorities), (risk of) malicious use, auditability, accountability (in case of harm), etc. The paper also elaborates on safety controls and guardrails, including documentation (datasheets for datasets), rate limiting, safety and content filters, detection models, hardcoding responses, watermarking, and community efforts (e.g. bias bounties). AI *policies* provide high-level guidelines and rules for organizational GenAI governance. They establish key objectives, accountability, roles, and responsibilities. Policies may mitigate risks but also limit beneficial uses (Solaiman, 2023). Model targets such as interpretability can be at odds with performance. Thus, guidance by internal risk policies defining how to evaluate and balance opposing objectives is needed. An analysis of 160 policies and guidelines across industries has yielded that policy statements typically balance innovation, ethical accountability and equitable access (Jiao et al., 2025). *Contractual* and legal measures such as licenses allow model providers or hosts a lever of control when users employ a system in non-desirable ways. For example, platforms hosting foundation models such as Huggingface use the Responsible AI License (RAIL) placing behavioral use conditions on a model. The model owner owns the license and bears responsibility for pursuing enforcement if needed (Contractor et al., 2020). Licenses are difficult to enforce for downloadable systems, as model behavior and uses are hard to monitor. *Compliance monitoring* ensures legal conformance and obedience to organizational policies, standards, procedures, and contractual agreements. For GenAI constant monitoring, e.g., using separate models such as OpenAI's Moderation API possibly supported by human oversight is important since, e.g., ethical compliance is difficult to guarantee. Human oversight can be distributed across different stages of GenAI-supported workflows, including drafting, refining, and reviewing (Taş et al.,2026). *Issue management* refers to identifying, managing, and resolving AI-related issues. Issues can stem from regulators that might enforce constraints that should be swiftly resolved. For example, Italy required OpenAI to adjust its privacy policy (McCallum, 2023).

Research provides best practices related to various *performance* aspects and *data and model targets*. For example, data protection (Li et al., 2025), data creation and valuation in general (Liang et al., 2022), data life cycle management (Zhang et al., 2023), and prompt engineering (White et al., 2023). Academic institutions have released policies on the appropriate use of GenAI by teachers and students that might be of value in an organizational setting. (Munoz et al., 2023) covers ethical behavior, attribution of content of GenAI, checking of outputs, as well as the development of AI literacy skills. Some works have discussed ethical principles for use, design and governance specific to their discipline, e.g., Harrer (2023) proposes for the medical and healthcare domain principles along the dimensions of human autonomy, well-being, XAI, accountability, equity and responsiveness. Other works prescribe ethical concepts across disciplines that might inform corporate ethics policies (Hagendorff, 2024). Furthermore, guidance for implementing AI governance practices within boards of directors has also been proposed (van Giffen & Ludwig, 2023). For AI regulation (Kahdan et al., 2023) describes four strategic response options (circumvent, comply, compromise, control).

*Scope monitoring* acknowledges the dynamics of antecedents including technological and legal evolution. As antecedents change, the scope also changes. Targets and governance must be reassessed and potentially updated leading to adaptive GenAI governance, e.g., through the use of structured sandboxes to learn about GenAI capabilities (Te'eni et al., 2026). Adaptive AI governance is fast, dynamic, involving many actors (Reuel & Undheim, 2024) as people, policies, systems, data, technology, processes, and other elements co-evolve (Janssen, 2025) in addition to evolution of ethical principles (Laine et al., 2025). Rapid evolution of technology can lower or enhance risks and enable new applications. For example, ChatGPT 3.5 scored about 70% on the MMLU benchmark covering a broad range of tasks in various domains(Hendrycks et al., 2021), while today's models achieve over 90% and come with strong reasoning capabilities (Schneider, 2025), increasingly allowing to execute workflows. As models quickly evolve and might be retired after less than a year (OpenAI Platform Documentation, 2025), processes must be in place to identify impact on existing deployments, e.g., re-evaluation of existing prompts designed by end-users or developers to assess if newer models yield benefits and prompts still work as expected. To account for rapid changes while maintaining some stability, establishing governance rules with different abstraction levels has been proposed to coordinate human-AI ecologies (Grisold et al., 2025). That is, more abstract rules covering fundamental values should be stable, while more specific coordination rules are changing more frequently.

**Relational mechanisms** support collaboration among stakeholders by incorporating communication, training, and coordination in decision-making. Human–AI collaboration governance requires leadership that frames AI's role, and clear human authorship (Huzooree, 2026). Social transparency (Sun et al., 2022) in AI systems makes socio-organizational factors visible that impact AI usage by presenting other users' interactions with the AI. "Learning from others" helps in understanding effective usage of AI and its outputs. In a study analyzing GenAI for software engineering (Sun et al., 2022), participants were interested in understanding how others evaluated model performance and their judgment of AI outputs, i.e., code quality, but also on how they generated outputs, i.e., code as well as strengths, weaknesses, common pitfalls, and limitations they experienced. Academic research has yet to provide detailed guidance and theorize on how to integrate GenAI in teams to coordinate decision-making (Brown et al., 2024). So far few works exist, e.g., for creative and routine tasks (Brynjolfsson et al., 2025; Han et al., 2024). For small and medium enterprises, collaboration with third parties such as research institutions has been recommended to handle resource constraints and obtain expertise (Rajaram & Tinguely, 2024). Technical mechanisms are concrete controls implemented in soft- or hardware to achieve targets, e.g., for *artifacts* such as data, models, and system. They often complement other mechanisms. For example, data quality targets can be reached by using algorithms for data cleaning (technical mechanism) but also providing standards and processes for data labeling for human annotators (procedural mechanism). Only few works, e.g., (Li et al., 2025; Scriven et al., 2023), discuss the interlinking of technical mechanisms in relation to governance. That is, practitioners have provided often rather narrow but detailed recommendations on applied technical governance, e.g., to combat weaknesses such as hallucinations of LLMs (Scriven et al., 2023). In contrast, tens of thousands of academic papers at top computer science conferences aim to improve specific data, model, and system targets, e.g., there were more than 21k submissions to NeurIPS in 2025 (Chairs, 2025). This makes a detailed treatment out-of-scope of this paper. But many technical mechanisms are not fully automatic and invisible to their users, but *human-facing* with a socio-technical component such as algorithms for explainability (Schneider, 2024). Thus, they might also demand non-technical governance mechanisms to ensure that humans and technology interact effectively. For example, evaluating LLMs is highly important but challenging leading to a number of benchmarks (Cao et al., 2025) but also auditing frameworks, e.g., the LLMAuditor(Amirizaniani et al., 2024) is a framework requiring a human-in-the-loop to identify model issues such as biases and inconsistencies. Still, fully automatic LLM-as-a-judge evaluations and more generic capability-based rather than task-specific evaluations are increasing(Cao et al., 2025). Research has also discussed how technology can *enable governance* itself, e.g., its operationalization. A concrete linking of governance expectations to engineering practice has been performed in (Raza et al., 2025). The paper proposes concrete metrics to operationalize targets, e.g., measures for fairness, but also assesses existing testbeds, benchmarks, metrics and documentation. They point towards gaps such as a lack of provenance and privacy benchmarks and practices as well as inconsistencies between metrics and documentation. Operationalizing AI governance is challenging and while desiderata have been proposed for tools, significant effort, e.g., to implement specific metrics for monitoring models and data properties, might be needed (Do et al., 2025).

### *Consequences*

Effective GenAI governance contributes to company objectives in several ways. *Performance effects* should be observable depending on the types of GenAI usages, e.g., efficiency gains of employees (S. Noy & Zhang, 2023), and increased customer satisfaction. *Risk management* should also benefit, e.g., in terms of reduced IP leakage risks, reduced reputational risks, and legal risks. Furthermore, *people* are impacted by governance. We conjecture that users and developers will have a positive attitude towards governance and the company as a whole, e.g., if the company aims to improve employees' capabilities through training, ensures that adequate corporate values related to GenAI are established. We also believe that *coordination* supported through governance can facilitate GenAI adoption within the company, as it reduces uncertainty within and outside the company.

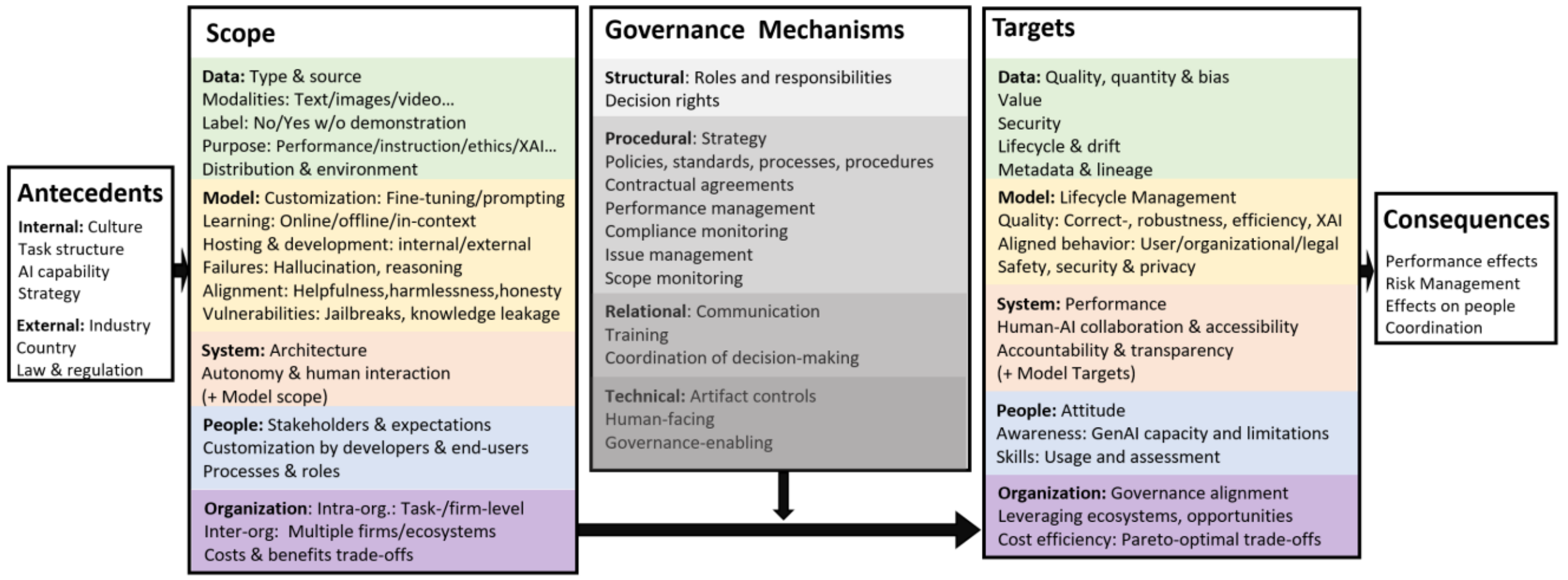

Figure 4. GenAI governance framework for organizations

## Discussion & Conclusion

Figure 4 summarizes our GenAI governance framework intended to support organizations in leveraging the value of GenAI while mitigating its risks. On a high level, it is similar to the AI risk management framework by (NIST, 2023;NIST, 2024) discussing mapping (scope), managing (mechanism), and measuring (targets) but focuses primarily on risks rather than other company-relevant aspects such as alignment with business objectives and value delivery. In response to the research question posed, our framework identifies the unique properties of GenAI and elaborates on the relevant dimensions of its governance. It thereby provides a gap-filling conceptualization that extends prior non-GenAI governance frameworks such as (Schneider et al., 2022). We argue that GenAI is (i) bottom-up, i.e., deployment and governance might be driven by employees and company departments (ii) hard to control technically because of the characteristics of foundation models, (iii) involving tasks of increasing complexity, and (iv) broadening the range of stakeholders who play a critical role in shaping AI behavior, leading to a novel "people" dimension in our framework. We specifically elaborated on the increased relevance of end-users. Prompting (in combination with task-specific data) effectively turns end-users into developers. End-users thus gain relevance and capability in leveraging individual GenAI opportunities, while also bearing more responsibility for mitigating risks like erroneous or harmful outputs. Organizational GenAI governance must account for this changing role of end-users. Likewise, it must account for the increasing technical complexity introduced by the growing scale of GenAI models, more complex training procedures and multimodal data. Compared to pre-GenAI frameworks, this considerably extends the data and model governance scope and introduces novel governance targets, as well as shifting prioritization of established targets. While companies might want to outsource GenAI development and hosting to external providers to sidestep technical challenges, this introduces its own governance challenges that must be adequately managed (e.g. through contracts). Additional governance challenges relate to the rapid pace of technical developments and the legal uncertainties of an evolving regulatory landscape. In response, GenAI governance approaches must be flexible, and companies must monitor advancements in technology and regulation. For example, companies might identify a need and assess existing solutions for detection mechanisms to spot GenAI generated artifacts (Knott et al., 2023).

Beyond extending prior governance frameworks, we theorize GenAI governance scope as endogenous to use. Conventional AI governance largely assumes that the governed artifact and its intended use can be specified ex ante. With GenAI, users can materially alter behavior, purpose, and associated risks through prompting, contextual data, and customization without changing the underlying model. Thus, use does not merely occur within a predefined governance scope; it can partly constitute that scope. This explains why people and scope monitoring become particularly important in GenAI governance.

With respect to the limitations of our framework, we recognize that future developments in technology and research might mitigate or intensify some of the issues here discussed. This could alter the governance scope and priorities proposed by our framework, emphasizing the importance of periodic review. For example, next-generation GenAI might be better in providing "uncertainty estimates" and answer with "I don't know", reducing the occurrence and associated risks of hallucinations. Another limitation stems from the methodology of a literature review, which depends on published studies and may not capture the most current or diverse perspectives in the field. However, to mitigate respective concerns, we included articles from arXiv.org in our review after rigorous quality assessment, acknowledging the rapid progress in GenAI research. Lastly, given the scope of this article, it does not detail the operationalization of the GenAI governance framework in organizational practice or empirical research. Respective knowledge would enhance its value, and we encourage future research to explore these aspects.

Our work has highlighted that open questions regarding effectively leveraging the value and mitigating the risk of GenAI remain. Thus, much needs to be done and it should be done quickly as GenAI is already in use by millions if not billions of people. We call for more works that aim to investigate governance from a business-oriented perspective, detailing and exploring governance mechanisms especially when it comes to people and organizational aspects, which we see as under-researched. Practitioners can leverage our framework to structure and validate their governance efforts. We hope that this leads to a well-balanced trade-off between the risks of GenAI shortcomings like hallucinations and legal concerns, and the opportunities to increase revenues and service quality, while accounting for ethical concerns – ultimately benefitting society.

**Disclosure statement:** The authors declare that they have no competing interests.

**Funding:** No funding received

**Data availability statement**:The publications analyzed in the review are identified in the reference list.

**Ethics statement:** Not required.

**Declaration of generative AI** and AI-assisted technologies in the manuscript preparation process: ChatGPT 5 was used mostly for feedback including polishing formulations. But no full passages were written by AI. Also, none of the dimensions or sub-dimensions stem from AI. Less than 5 references stem from AI prior to 2026. The literature search in the last iteration conducted in 2026 evolved also the use of ChatGPT 5.6 Pro and Extra High in addition to Google Scholar.

## References

Abraham, R., Schneider, J., & Vom Brocke, J. (2019). Data governance: A conceptual framework, structured review, and research agenda. *International Journal of Information Management*.

Alshammari, K. H. (2026). Generative AI, epistemic authority, and knowledge governance tension: implications for innovation novelty and reliability. *Future Business Journal*, *12*(1), 222.

Amirizaniani, M., Yao, J., Lavergne, A., Okada, E. S., Chadha, A., Roosta, T., & Shah, C. (2024). LLMAuditor: A Framework for Auditing Large Language Models Using Human-in-the-Loop. *arXiv Preprint arXiv:2402.09346*.

Annapureddy, R., Fornaroli, A., & Gatica-Perez, D. (2024). Generative AI literacy: Twelve defining competencies. *Digital Government: Research and Practice*.

Askell, A., Bai, Y., Chen, A., Drain, D., Ganguli, D., Henighan, T., Jones, A., Joseph, N., Mann, B., DasSarma, N., & others. (2021). A general language assistant as a laboratory for alignment. *arXiv Preprint arXiv:2112.00861*.

Ayers, J. W., Poliak, A., Dredze, M., Leas, E. C., Zhu, Z., Kelley, J. B., Faix, D. J., Goodman, A. M., Longhurst, C. A., Hogarth, M., & others. (2023). Comparing physician and artificial intelligence chatbot responses to patient questions posted to a public social media forum. *JAMA Internal Medicine*.

Banh, L., & Strobel, G. (2023). Generative artificial intelligence. *Electronic Markets*.

Becker, J., Rush, N., Barnes, E., & Rein, D. (2025). Measuring the impact of early-2025 AI on experienced open-source developer productivity. *arXiv Preprint arXiv:2507.09089*.

Bell, G., Burgess, J., Thomas, J., & Sadiq, S. (2023). *Generative AI - Language Models (LLMs) and Multimodal Foundation Models (MFMs)*. Australian Council of Learned Academies.

Birkstedt, T., Minkkinen, M., Tandon, A., & Mäntymäki, M. (2023). AI governance: Themes, knowledge gaps and future agendas. *Internet Research*.

Bommasani, R., Hudson, D. A., Adeli, E., Altman, R., Arora, S., von Arx, S., Bernstein, M. S., Bohg, J., Bosselut, A., Brunskill, E., & others. (2021). On the opportunities and risks of foundation models. *arXiv Preprint arXiv:2108.07258*.

Brodsky, S. (2024). *The hidden costs of AI: How generative models are reshaping corporate budgets*. IBM. https://www.ibm.com/think/insights/ai-economics-compute-cost

Brown, O., Davison, R. M., Decker, S., Ellis, D. A., Faulconbridge, J., Gore, J., Greenwood, M., Islam, G., Lubinski, C., MacKenzie, N. G., & others. (2024). Theory-driven perspectives on generative artificial intelligence in business and management. *British Journal of Management*, *35*(1), 3–23.

Brown, T., Mann, B., Ryder, N., Subbiah, M., Kaplan, J. D., Dhariwal, P., Neelakantan, A., Shyam, P., Sastry, G., Askell, A., & others. (2020). Language models are few-shot learners. *Advances in Neural Information Processing Systems*, *33*, 1877–1901.

Brynjolfsson, E., Li, D., & Raymond, L. R. (2025). “Generative AI at Work.” *The Quarterly Journal of Economics, 140*(2), 889–942.

Calvino, F., Reijerink, J., & Samek, L. (2025). The effects of generative AI on productivity, innovation and entrepreneurship. *OECD Artificial Intelligence Papers*.

Cao, Y., Hong, S., Li, X., Ying, J., Ma, Y., Liang, H., Liu, Y., Yao, Z., Wang, X., Huang, D., & others. (2025). Toward generalizable evaluation in the llm era: A survey beyond benchmarks. *arXiv Preprint arXiv:2504.18838*.

Carlini, N., Ippolito, D., Jagielski, M., Lee, K., Tramer, F., & Zhang, C. (2022). Quantifying Memorization Across Neural Language Models. *International Conference on Learning Representations*.

Celikyilmaz, A., Clark, E., & Gao, J. (2020). Evaluation of text generation: A survey. *arXiv Preprint arXiv:2006.14799*.

Chairs, C. (2025). *Reflections on the 2025 Review Process from the Program Committee Chairs*. https://blog.neurips.cc/2025/09/30/reflections-on-the-2025-review-process-from-the-program-committee-chairs/

Chowdhury, S., Budhwar, P., & Wood, G. (2024). Generative artificial intelligence in business: Towards a strategic human resource management framework. *British Journal of Management*.

Contractor, D., McDuff, D., Haines, J. K., Lee, J., Hines, C., & Hecht, B. J. (2020). Behavioral Use Licensing for Responsible AI. *arXiv Preprint arXiv:2011.03116*.

Dencik, J., Goehring, B., & Marshall, A. (2023). Managing the emerging role of generative AI in next-generation business. *Strategy & Leadership*.

Deng, Y., Zhang, W., Pan, S. J., & Bing, L. (2023). Multilingual jailbreak challenges in large language models. *arXiv Preprint arXiv:2310.06474*.

Diaferia, L., Kappelhoff, S. L., & Blohm, I. (2026). From generic to strategic: Managerial challenges and response mechanisms for customizing Gen AI in companies. In *Proceedings of the 59th Hawaii International Conference on System Sciences*.

Do, H. J., Babbar, S., Li, W., Walks, L., & Misko, S. (2025). Understanding Industry Practitioners' Experiences in Generative AI Governance. *Proceedings of the Extended Abstracts of the CHI Conference on Human Factors in Computing Systems*, 1–8.

Dong, Y., Mu, R., Zhang, Y., Sun, S., Zhang, T., Wu, C., Jin, G., Qi, Y., Hu, J., Meng, J., & others. (2024). Safeguarding Large Language Models: A Survey. *arXiv Preprint arXiv:2406.02622*.

Ebell, C., Baeza-Yates, R., Benjamins, R., Cai, H., Coeckelbergh, M., Duarte, T., Hickok, M., Jacquet, A., Kim, A., Krijger, J., Macintyre, J., Madhamshettiwar, P., Maffeo, L., Matthews, J., Medsker, L., Smith, P., & Thais, S. (2021). Towards Intellectual Freedom in an AI Ethics Global Community. *AI and Ethics*.

El Rassi, M. A. B. (2026). Speed with fit: Gen-AI, governance, and the transformation of creative work under crisis. In *ECIS 2026 proceedings*.

Emett, S., Eulerich, M., Pikoos, J., & Wood, D. A. (2026). The development of a generative artificial intelligence (AI) governance framework. *Accounting Horizons*, 1-16.

Esposito De Falco, S., Laviola, F., Mercuri, F., & Cucari, N. (2026). Different routes, same storm: a three-dimensional paradox view of generative AI's governance. *Management Decision*, 1-27.

EU. (2024). EU AI Act. *Official Journal of the European Union*, L 2024/1689.

Ferraro, C., Demsar, V., Sands, S., Restrepo, M., & Campbell, C. (2024). The paradoxes of generative AI-enabled customer service: A guide for managers. *Business Horizons*, *67*(5), 549–559.

Fui-Hoon Nah, F., Zheng, R., Cai, J., Siau, K., & Chen, L. (2023). Generative AI and ChatGPT: Applications, challenges, and AI-human collaboration. *Journal of Information Technology Case and Application Research*.

Furqan, T., Aleissaee, H. M., Ghenimi, N., & Arabi, A. A. (2026). Global laws governing intellectual property rights for AI-generated works. *Discover Artificial Intelligence*.

Gillespie, N., Lockey, S., Ward, T., Macdade, A., & Hassed, G. (2025). *Trust, attitudes and use of artificial intelligence: A global study 2025*. The University of Melbourne and KPMG

Grisold, T., Berente, N., & Seidel, S. (2025). Guardrails for human-AI ecologies: Norm-based coordination and design for predictability. *MIS Quarterly*, *49*(4), 1239–1266.

Grynbaum, M. M., & Mac, R. (2023). *The Times Sues OpenAI and Microsoft Over A.I. Use of Copyrighted Work*. https://www.nytimes.com/2023/12/27/business/media/new-york-times-open-ai-microsoft-lawsuit.html

Guldimann, P., Spiridonov, A., Staab, R., Jovanović, N., Vero, M., Vechev, V., ... & Vechev, M. (2024). Compl-ai framework: A technical interpretation and llm benchmarking suite for the eu artificial intelligence act. *arXiv preprint arXiv:2410.07959*.

Gupta, M., Akiri, C., Aryal, K., Parker, E., & Praharaj, L. (2023). From ChatGPT to ThreatGPT: Impact of generative AI in cybersecurity and privacy. *IEEE Access*.

Hagendorff, T. (2024). Mapping the Ethics of Generative AI: A Comprehensive Scoping Review. *arXiv Preprint arXiv:2402.08323*.

Hai, S., Long, T., Honora, A., Japutra, A., & Guo, T. (2025). The dark side of employee-generative AI collaboration in the workplace: An investigation on work alienation and employee expediency. *International Journal of Information Management*, *83*, 102905.

Han, Y., Qiu, Z., Cheng, J., & LC, R. (2024). When Teams Embrace AI: Human Collaboration Strategies in Generative Prompting in a Creative Design Task. *Proceedings of the CHI Conference on Human Factors in Computing Systems*.

Harrer, S. (2023). Attention is not all you need: The complicated case of ethically using large language models in healthcare and medicine. *EBioMedicine*, *90*.

Hendrycks, D., Burns, C., Basart, S., Zou, A., Mazeika, M., Song, D., & Steinhardt, J. (2021). Measuring Massive Multitask Language Understanding. *International Conference on Learning Representations (ICLR) 2021 Workshop Track*. https://arxiv.org/abs/2009.03300

Henry, K. E., Kornfield, R., Sridharan, A., Linton, R. C., Groh, C., Wang, T., Wu, A., Mutlu, B., & Saria, S. (2022). Human–machine teaming is key to AI adoption: Clinicians' experiences with a deployed machine learning system. *NPJ Digital Medicine*, *5*(1), 97.

Horneber, D., & Laumer, S. (2023). Algorithmic Accountability. *Business & Information Systems Engineering*.

Huang, Y., Arora, C., Houng, W. C., Kanij, T., Madulgalla, A., & Grundy, J. (2025). Ethical concerns of generative AI and mitigation strategies: A systematic mapping study. *arXiv Preprint arXiv:2502.00015*.

Hughes, L., Davies, F., Li, K., Gunaratnege, S. M., Malik, T., & Dwivedi, Y. K. (2026). Beyond the hype: Organisational adoption of Generative AI through the lens of the TOE framework–A mixed methods perspective. *International journal of information management*, *86*, 102982.

Hupont, I., Micheli, M., Delipetrev, B., Gómez, E., & Garrido, J. S. (2023). Documenting high-risk AI: a European regulatory perspective. *Computer*.

Huzooree, G. (2026a). E-leadership and human-AI collaboration: Socio-technical alignment in project-based teams. *Journal of Organizational Effectiveness: People and Performance*.

ISO/IEC. (2023). *Information technology — Artificial intelligence — Management system* (ISO/IEC 42001:2023). https://www.iso.org/standard/81230.html

Janssen, M. (2025). Responsible governance of generative AI: conceptualizing GenAI as complex adaptive systems. *Policy and Society*, *44*(1), 38–51.

Jernite, Y., Nguyen, H., Biderman, S., Rogers, A., Masoud, M., Danchev, V., Tan, S., Luccioni, A. S., Subramani, N., Johnson, I., & others. (2022). Data governance in the age of large-scale data-driven language technology. *Proceedings of the 2022 ACM Conference on Fairness, Accountability, and Transparency*, 2206–2222.

Jiao, J., Afroogh, S., Chen, K., Atkinson, D., & Dhurandhar, A. (2025). Generative AI and LLMs in Industry: A text-mining Analysis and Critical Evaluation of Guidelines and Policy Statements Across Fourteen Industrial Sectors. *arXiv Preprint arXiv:2501.00957*.

Joosten, J., Bilgram, V., Hahn, A., & Totzek, D. (2024). Comparing the Ideation Quality of Humans With Generative Artificial Intelligence. *IEEE Engineering Management Review*.

Kahdan, M., Hartwich, N. J., Salge, O., & Cichy, P. (2023). Navigating Uncertain Waters: How Organizations Respond to Institutional Pressure in Times of the Looming EU AI Act. *International Conference on Information Systems (ICIS)*.

Kalai, A. T., Nachum, O., Vempala, S. S., & Zhang, E. (2025). Why language models hallucinate. *arXiv preprint arXiv:2509.04664*.

Khurana, A., Subramonyam, H., & Chilana, P. K. (2024). Why and When LLM-Based Assistants Can Go Wrong: Investigating the Effectiveness of Prompt-Based Interactions for Software Help-Seeking. *Proceedings of the 29th International Conference on Intelligent User Interfaces*, 288–303.

King, W. R., & He, J. (2005). Understanding the role and methods of meta-analysis in IS research. *Communications of the Association for Information Systems*, *16*(1), 32.

Knott, A., Pedreschi, D., Chatila, R., Chakraborti, T., Leavy, S., Baeza-Yates, R., Eyers, D., Trotman, A., Teal, P. D., Biecek, P., & others. (2023). Generative AI models should include detection mechanisms as a condition for public release. *Ethics and Information Technology*.

Kojima, T., Gu, S. S., Reid, M., Matsuo, Y., & Iwasawa, Y. (2022). Large Language Models are Zero-Shot Reasoners. *arXiv Preprint arXiv:2205.11916*.

Koplin, J. J. (2023). Dual-use implications of AI text generation. *Ethics and Information Technology*.

Laine, J., Minkkinen, M., & Mäntymäki, M. (2025). Understanding the ethics of generative AI: Established and new ethical principles. *Communications of the Association for Information Systems*, *56*(1), 7.

Lee, H.-P., Sarkar, A., Tankelevitch, L., Drosos, I., Rintel, S., Banks, R., & Wilson, N. (2025). The impact of generative AI on critical thinking: Self-reported reductions in cognitive effort and confidence effects from a survey of knowledge workers. In *Proceedings of the 2025 CHI Conference on Human Factors in Computing Systems*

Lee, K., Ippolito, D., Nystrom, A., Zhang, C., Eck, D., Callison-Burch, C., & Carlini, N. (2021). Deduplicating training data makes language models better. *arXiv Preprint arXiv:2107.06499*.

Leidner, D. E. (2018). Review and theory symbiosis: An introspective retrospective. *Journal of the Association for Information Systems*, *19*(6), 1.

Leon, M. (2026). Governing generative AI in organizations: a design theory and quasi-experimental field study of sociotechnical guardrails: M. Leon. *The Journal of Supercomputing*, *82*(12), 641.

Lewis, P., Perez, E., Piktus, A., Petroni, F., Karpukhin, V., Goyal, N., Küttler, H., Lewis, M., Yih, W., Rocktäschel, T., & others. (2020). Retrieval-augmented generation for knowledge-intensive nlp tasks. *Advances in Neural Information Processing Systems*.

Li, K., Zhang, F., Hughes, L., & Griffin, M. A. (2026). Leveraging generative AI for project management: The role of mindfulness and job crafting. *International Journal of Project Management, 44*(2),

Li, Y., Shao, S., He, Y., Guo, J., Zhang, T., Qin, Z., Chen, P.-Y., Backes, M., Torr, P., Tao, D., & others. (2025). Rethinking data protection in the (generative) artificial intelligence era. *arXiv Preprint arXiv:2507.03034*.

Liang, W., Tadesse, G. A., Ho, D., Fei-Fei, L., Zaharia, M., Zhang, C., & Zou, J. (2022). Advances, challenges and opportunities in creating data for trustworthy AI. *Nature Machine Intelligence*, *4*(8), 669–677.

Lin, J., Yu, Y., Zhou, Y., Zhou, Z., & Shi, X. (2020). How many preprints have actually been printed and why: A case study of computer science preprints on arXiv. *Scientometrics*, *124*(1), 555–574.

Liu, P., Yuan, W., Fu, J., Jiang, Z., Hayashi, H., & Neubig, G. (2023). Pre-train, prompt, and predict: A systematic survey of prompting methods in natural language processing. *ACM Computing Surveys*, *55*(9), 1–35.

Liu, Y., Zhang, K., Li, Y., Yan, Z., Gao, C., Chen, R., Yuan, Z., Huang, Y., Sun, H., Gao, J., & others. (2024). Sora: A Review on Background, Technology, Limitations, and Opportunities of Large Vision Models. *arXiv Preprint arXiv:2402.17177*.

Longo, L., Brcic, M., Cabitza, F., Choi, J., Confalonieri, R., Ser, J. D., Guidotti, R., Hayashi, Y., Herrera, F., Holzinger, A., & others. (2024). Explainable artificial intelligence (XAI) 2.0: A manifesto of open challenges and interdisciplinary research directions. *Information Fusion*.

Longpre, S., Mahari, R., Chen, A., et al. (2024). "A large-scale audit of dataset licensing and attribution in AI." *Nature Machine Intelligence, 6*, 975–987.

Lu, Y., Bartolo, M., Moore, A., Riedel, S., & Stenetorp, P. (2021). Fantastically ordered prompts and where to find them: Overcoming few-shot prompt order sensitivity. *arXiv Preprint arXiv:2104.08786*.

Maas, M. M. (2023). Advanced AI Governance: A Literature Review of Problems, Options, and Proposals. *AI Foundations Report*, *4*.

Madaan, A., Zhou, S., Alon, U., Yang, Y., & Neubig, G. (2022). Language Models of Code are Few-Shot Commonsense Learners. *Conference on Empirical Methods in Natural Language Processing*.

Mäntymäki, M., Minkkinen, M., Birkstedt, T., & Viljanen, M. (2022). Putting AI ethics into practice: The hourglass model of organizational AI governance. *arXiv Preprint arXiv:2206.00335*.

Martín-Martín, A., Thelwall, M., Orduna-Malea, E., & Delgado López-Cózar, E. (2021). Google Scholar, Microsoft Academic, Scopus, Dimensions, Web of Science, and OpenCitations' COCI: a multidisciplinary comparison of coverage via citations. *Scientometrics*, *126*(1), 871–906.

Matt Gean. (2023). *Artificial Intelligence CEOs Warning*. https://edition.cnn.com/2023/06/14/business/artificial-intelligence-ceos-warning/index.html

McAfee, A., Rock, D., & Brynjolfsson, E. (2023). How to Capitalize on Generative AI. *Harvard Business Review*. https://hbr.org/2023/11/how-to-capitalize-on-generative-ai

McCallum, S. (2023). *ChatGPT banned in Italy over privacy concerns*. https://www.bbc.com/news/technology-65139406

Memmert, L., & Tavanapour, N. (2023). Towards Human-AI-Collaboration in Brainstorming: Empirical Insights into the Perception of Working with a Generative AI. *European Conference on Information Systems (ECIS)*.

McKinsey. (2025). *The state of AI: How organizations are rewiring to capture value*. [https://www.mckinsey.com/capabilities/quantumblack/our-insights/the-state-of-ai-how-organizations-are-rewiring-to-capture-value](https://www.mckinsey.com/capabilities/quantumblack/our-insights/the-state-of-ai-how-organizations-are-rewiring-to-capture-value)

Meta. (2024). *Meta LLaMA-3*. https://ai.meta.com/blog/meta-llama-3/

Meta AI. (2025). *LLaMA 4: Multimodal Intelligence*. Meta Platforms, Inc. https://ai.meta.com/blog/llama-4-multimodal-intelligence/

Microsoft. (2023). *Get Started with Copilot in Microsoft Teams Meetings*. https://support.microsoft.com/en-au/office/get-started-with-copilot-in-microsoft-teams-meetings-0bf9dd3c-96f7-44e2-8bb8-790bedf066b1

Microsoft. (2024). *Overview of Copilot for Power BI (preview)*. https://learn.microsoft.com/en-us/power-bi/create-reports/copilot-introduction

Mora-Cantallops, M. & others. (2021). Traceability for Trustworthy AI: A Review of Models and Tools. *Big Data and Cognitive Computing*, *5*(2).

Munoz, A., Wilson, A., Pereira Nunes, B., Del Medico, C., Slade, C., Bennett, D., Tyler, D., Seymour, E., Hepplewhite, G., Randell-Moon, H., & others. (2023). *AAIN Generative Artificial Intelligence Guidelines*.

Murray, M. D. (2023). Generative and AI Authored Artworks and Copyright Law. *Hastings Communications & Entertainment Law Journal*.

Nasr, M., Carlini, N., Hayase, J., Jagielski, M., Cooper, A. F., Ippolito, D., Choquette-Choo, C. A., Wallace, E., Tramèr, F., & Lee, K. (2023). Scalable extraction of training data from (production) language models. *arXiv Preprint arXiv:2311.17035*.

NATO. (2021). *Summary of the NATO Artificial Intelligence Strategy*. https://www.nato.int/cps/en/natohq/official_texts_187617.htm

Newman, J. (2023). *A Taxonomy of Trustworthiness for Artificial Intelligence*. Center for Long-Term Cybersecurity (CLTC), Berkeley. https://cltc.berkeley.edu/publication/a-taxonomy-of-trustworthiness-for-artificial-intelligence/

Nguyen, P., Watson, G. P., Barnes, D., Agrawal, S., Schuster, A. M., & Cotten, S. R. (2026). Navigating workplace AI adoption: The influence of perceptions and affective attitudes on employees' intentions to use AI at work. *Journal of Management & Organization*, *32*(4), 1018-1043.

Nicholson Price II, W. (2022). *Risks and Remedies for Artificial Intelligence in Health Care*. https://www.brookings.edu/research/risks-and-remedies-for-artificial-intelligence-in-health-care/

Nickerson, R. C., Varshney, U., & Muntermann, J. (2013). A method for taxonomy development and its application in information systems. *European Journal of Information Systems*, *22*, 336–359.

NIST. (2023). *Artificial Intelligence Risk Management Framework (AI RMF 1.0)*. https://doi.org/10.6028/NIST.AI.100-1

NIST. (2024). *Artificial Intelligence Risk Management Framework: Generative Artificial Intelligence Profile* (NIST AI 600-1). https://doi.org/10.6028/NIST.AI.600-1

Noy, S., & Zhang, W. (2023). Experimental evidence on the productivity effects of generative artificial intelligence. *Science*.

Okoli, C., & Schabram, K. (2015). *A guide to conducting a systematic literature review of information systems research*.

Oldehoeft, A. E. (1992). *Foundations of a Security Policy for Use of the National Research and Educational Network*.

Oniani, D., Hilsman, J., Peng, Y., Poropatich, R. K., Pamplin, C., Legault, L., Wang, Y., & others. (2023). From Military to Healthcare: Adopting and Expanding Ethical Principles for Generative Artificial Intelligence. *arXiv Preprint arXiv:2308.02448*.

OpenAI. (2023a). GPT-4 Technical Report. *arXiv Preprint arXiv:2303.08774*.

OpenAI. (2023c). *Introducing the GPT Store*. https://openai.com/blog/introducing-the-gpt-store

OpenAI. (2023d). *Overview of OpenAI Assistants*. https://platform.openai.com/docs/assistants/overview

OpenAI. (2024). *OpenAI's Privacy Policy*. https://openai.com/policies/privacy-policy

OpenAI. (2025). *GPT-5 System Card*. OpenAI. https://deploymentsafety.openai.com/gpt-5

OpenAI Platform Documentation. (2025). *Deprecations*. OpenAI. https://platform.openai.com/docs/deprecations

Ouyang, L., Wu, J., Jiang, X., Almeida, D., Wainwright, C., Mishkin, P., Zhang, C., Agarwal, S., Slama, K., Ray, A., & others. (2022). Training language models to follow instructions with human feedback. *Advances in Neural Information Processing Systems*, *35*, 27730–27744.

Plugge, A., Beulen, E., & van Hillegersberg, J. (2026). Exploring the governance of AI outsourcing from a client-service provider perspective: Realising business advantages. In *ECIS 2026 proceedings*.

Rajaram, K., & Tinguely, P. N. (2024). Generative artificial intelligence in small and medium enterprises: Navigating its promises and challenges. *Business Horizons*.

Raza, S., Qureshi, R., Zahid, A., Kamawal, S., Sadak, F., Fioresi, J., Saeed, M., Sapkota, R., Jain, A., Zafar, A., & others. (2025). Who is responsible? The data, models, users or regulations? A comprehensive survey on responsible generative ai for a sustainable future. *arXiv Preprint arXiv:2502.08650*.

Rebedea, T., Dinu, R., Sreedhar, M. N., Parisien, C., & Cohen, J. (2023). NeMo Guardrails: A Toolkit for Controllable and Safe LLM Applications with Programmable Rails. *Proceedings of the 2023 Conference on Empirical Methods in Natural Language Processing: System Demonstrations*, 431–445.

Reddy, S. (2024). Generative AI in healthcare: An implementation science informed translational path on application, integration and governance. *Implementation Science*, *19*(1), 27.

Reid, M., Savinov, N., Teplyashin, D., Lepikhin, D., Lillicrap, T., Alayrac, J., Soricut, R., Lazaridou, A., Firat, O., Schrittwieser, J., & others. (2024). Gemini 1.5: Unlocking multimodal understanding across millions of tokens of context. *arXiv Preprint arXiv:2403.05530*.

Reuel, A., & Undheim, T. A. (2024). Generative AI needs adaptive governance. *arXiv Preprint arXiv:2406.04554*.

Roht, K., Nikiforova, A., & Lefebvre, H. (2026). A typology for responsible delegation of data quality management to generative AI. In *ECIS 2026 proceedings*. Association for Information Systems.

Ruokonen, M., & Ritala, P. (2025). Managing Generative AI for Strategic Advantage. *Research-Technology Management*, *68*(4), 11–20.

Safaei, D., Hein, A., & Haki, K. (2026). Logic hybridization in enterprise platforms: Reconfiguring architecture and governance for generative AI. *Journal of Information Technology*. Advance online publication.

Sastry, G., Heim, L., Belfield, H., Anderljung, M., Brundage, M., Hazell, J., O'Keefe, C., Hadfield, G. K., Ngo, R., Pilz, K., & others. (2024). Computing Power and the Governance of Artificial Intelligence. *arXiv Preprint arXiv:2402.08797*.

Schäfer, M., Schneider, J., Drechsler, K., & vom Brocke, J. (2022). AI governance: are Chief AI Officers and AI Risk Officers needed? *European Conference of Information Systems (ECIS)*.

Schaal, S. (1996). Learning from demonstration. *Advances in Neural Information Processing Systems*.

Schick, T., Dwivedi-Yu, J., Dessì, R., Raileanu, R., Lomeli, M., Hambro, E., Zettlemoyer, L., Cancedda, N., & Scialom, T. (2024). Toolformer: Language models can teach themselves to use tools. *Advances in Neural Information Processing Systems*, *36*.

Schmidt, M., & Schöbel, S. (2026). From prompt to process: Qualitative insights on how GenAI use rewrites startup structures. In *ECIS 2026 proceedings*.

Schmitt, K., & Grauer, T. (2026). Generative artificial intelligence governance: A structural practice perspective. *ECIS 2026 Proceedings.*

Schneider, J. (2024). Explainable Generative AI (GenXAI): A Survey, Conceptualization, and Research Agenda. *arXiv Preprint arXiv:2404.09554*.

Schneider, J. (2025). Generative to Agentic AI: Survey, Conceptualization, and Challenges. *arXiv Preprint arXiv:2504.18875*. https://arxiv.org/abs/2504.18875

Schneider, J., Abraham, R., Meske, C., & Brocke, J. V. (2022). Artificial Intelligence Governance For Businesses. *Information Systems Management*.

Schneider, J., Meske, C., & Kuss, P. (2024b). Foundation Models. *Business Information Systems Engineering*.

Schneider, Johannes, Sinem, K., & Stockhammer, D. (2024a). Empowering Clients: Transformation of Design Processes Due to Generative AI. *arXiv Preprint arXiv:2411.15061*.

Schulhoff, Sander, Ilie, M., Balepur, N., Kahadze, K., Liu, A., Si, C., Li, Y., Gupta, A., Han, H., Schulhoff, Sevien, & others. (2024). The Prompt Report: A Systematic Survey of Prompting Techniques. *arXiv Preprint arXiv:2406.06608*.

Scott Likens and Nicole Wakefield. (2023). *Early days for generative AI strategy*.

Scriven, G., Braga, M., Santos, D., & Sarai, D. (2023). Applied Generative Ai Governance: A Viable Model Through Control Automation. *Journal of Financial Transformation*, *58*, 24–33.

Sheehan, M. (2023). China's AI regulations and how they get made. *Carnegie Endowment for International Peace*.

Sigfrids, A., Leikas, J., Salo-Pöntinen, H., & Koskimies, E. (2023). Human-centricity in AI governance: A systemic approach. *Frontiers in Artificial Intelligence*.

Simkute, A., Tankelevitch, L., Kewenig, V., Scott, A. E., Sellen, A., & Rintel, S. (2024). Ironies of Generative AI: Understanding and mitigating productivity loss in human-AI interactions. *arXiv Preprint arXiv:2402.11364*.

Solaiman, I. (2023). The gradient of generative AI release: Methods and considerations. *ACM Conference on Fairness, Accountability, and Transparency*.

Sposato, M. (2026). Shadow AI learning in organisations: a sociomaterial framework for employee-driven artificial intelligence training and knowledge management. *Journal of Knowledge Management*, 1-18.

Sui, Y., Zhou, Mengyu, Zhou, Mingjie, Han, S., & Zhang, D. (2024). Table meets llm: Can large language models understand structured table data? A benchmark and empirical study. *Proceedings of the 17th ACM International Conference on Web Search and Data Mining*, 645–654.

Sun, J., Liao, Q. V., Muller, M., Agarwal, M., Houde, S., Talamadupula, K., & Weisz, J. D. (2022). Investigating explainability of generative AI for code through scenario-based design. *Int. Conference on Intelligent User Interfaces*.

Symeonidis, D., Schneider, J., & Nikiforova, A. (2026). Toward cultural-ethical fit in AI governance: A global typology of governance profiles. *Proceedings of the 34th European Conference on Information Systems (ECIS)*.

Taeihagh, A. (2025). Governance of generative AI. In *Policy and society* (Vol. 44, Issue 1, pp. 1–22). Oxford University Press UK.

Tan, T. F., Thirunavukarasu, A. J., Campbell, J. P., Keane, P. A., Pasquale, L. R., Abramoff, M. D., Kalpathy-Cramer, J., Lum, F., Kim, J. E., Baxter, S. L., & others. (2023). Generative Artificial Intelligence Through ChatGPT and Other Large Language Models in Ophthalmology: Clinical Applications and Challenges. *Ophthalmology Science*, *3*(4).

Taş, E., Memmert, L., & Bittner, E. (2026). Episodic oversight in generative AI workflows: A nine-step protocol for preserving human agency (OP-9). *Electronic Markets*

Te'eni, D., Raymond, M., Rowe, F., Thénoz, E., & Trimborn, P. (2026). Organizational learning for exploring Generative AI: CORE-sandbox experiments. *International Journal of Information Management*, 103029.

Toorajipour, R., Perozzo, H., Ravarini, A., Eaton, B., & Ben Ajami, M. (2026). Navigating the generative AI adoption dilemma: Pathways and trade-offs through a data ecosystem perspective. In *Proceedings of the 59th Hawaii International Conference on System Sciences*

U.S. Copyright Office. (2025). *Copyright and artificial intelligence, Part 2: Copyrightability*. https://www.copyright.gov/ai/Copyright-and-Artificial-Intelligence-Part-2-Copyrightability-Report.pdf

van Giffen, B., & Ludwig, H. (2023). How Boards of Directors Govern Artificial Intelligence. *MIS Quarterly Executive*.

Van Veen, D., Van Uden, C., Blankemeier, L., Delbrouck, J.-B., Aali, A., Bluethgen, C., Pareek, A., Polacin, M., Reis, E. P., Seehofnerová, A., & others. (2024). Adapted large language models can outperform medical experts in clinical text summarization. *Nature Medicine*, 1–9.

Vaswani, A., Shazeer, N., Parmar, N., Uszkoreit, J., Jones, L., Gomez, A. N., Kaiser, \Lukasz, & Polosukhin, I. (2017). Attention is all you need. *Advances in Neural Information Processing Systems*, *30*.

Vats, V., Nizam, M. B., Liu, M., Wang, Z., Ho, R., Prasad, M. S., Titterton, V., Malreddy, S. V., Aggarwal, R., Xu, Y., & others. (2024). A Survey on Human-AI Teaming with Large Pre-Trained Models. *arXiv Preprint arXiv:2403.04931*.

von Brackel-Schmidt, C., Kučević, E., Memmert, L., Tavanapour, N., Cvetkovic, I., Bittner, E. A., & Böhmann, T. (2023). A User-centric Taxonomy for Conversational Generative Language Models. *International Conference on Information Systems (ICIS)*.

Wagner, G., Prester, J., Mousavi, R., Lukyanenko, R., & Paré, G. (2026). “Generative artificial intelligence for literature reviews.” *Journal of Information Technology, 41*(2), 148–170. DOI: 10.1177/02683962261425675.

Wamba, S. F., Queiroz, M. M., Randhawa, K., & Gupta, G. (2025). Generative artificial intelligence and the challenges to adding value ethically. *Technovation, 144*, 103235.

Weidinger, L., Rauh, M., Marchal, N., Manzini, A., Hendricks, L. A., Mateos-Garcia, J., Bergman, S., Kay, J., Griffin, C., Bariach, B., & others. (2023). Sociotechnical Safety Evaluation of Generative AI Systems. *arXiv Preprint arXiv:2310.11986*.

Weiser, B. (2023). Here’s What Happens When Your Lawyer Uses ChatGPT. *The New York Times*. https://www.nytimes.com/2023/05/27/nyregion/avianca-airline-lawsuit-chatgpt.html

White, J., Fu, Q., Hays, S., Sandborn, M., Olea, C., Gilbert, H., Elnashar, A., Spencer-Smith, J., & Schmidt, D. C. (2023). A prompt pattern catalog to enhance prompt engineering with chatgpt. *arXiv Preprint arXiv:2302.11382*.

Wodecki, B. (2024). *Meet Alpaca: The Open Source ChatGPT Made for Less Than $600*. https://aibusiness.com/nlp/meet-alpaca-the-open-source-chatgpt-made-for-less-than-600

Xu, Z., Liu, Z., Chen, B., Tang, Y., Wang, J., Zhou, K., Hu, X., & Shrivastava, A. (2023). Compress, then prompt: Improving accuracy-efficiency trade-off of llm inference with transferable prompt. *arXiv Preprint arXiv:2305.11186*.

Yang, J., Deng, W., Liu, B., Huang, Y., & Li, X. (2023). Matching-based Data Valuation for Generative Model. *arXiv Preprint arXiv:2304.10701*.

Zamfirescu-Pereira, J., Wong, R. Y., Hartmann, B., & Yang, Q. (2023). Why Johnny can’t prompt: How non-AI experts try (and fail) to design LLM prompts. *Proceedings of the 2023 CHI Conference on Human Factors in Computing Systems*, 1–21.

Zhang, D., Xia, B., Liu, Y., Xu, X., Hoang, T., Xing, Z., Staples, M., Lu, Q., & Zhu, L. (2023). Navigating privacy and copyright challenges across the data lifecycle of generative ai. *arXiv Preprint arXiv:2311.18252*.

Zhang, H., Yu, Y., Jiao, J., Xing, E., El Ghaoui, L., & Jordan, M. (2019). Theoretically principled trade-off between robustness and accuracy. *International Conference on Machine Learning*.

Zhou, C., Liu, P., Xu, P., Iyer, S., Sun, J., Mao, Y., Ma, X., Efrat, A., Yu, P., Yu, L., & others. (2024). Lima: Less is more for alignment. *Advances in Neural Information Processing Systems*, *36*.

Zou, W., Geng, R., Wang, B., & Jia, J. (2025). "PoisonedRAG: Knowledge Corruption Attacks to Retrieval-Augmented Generation of Large Language Models." *34th USENIX Security Symposium*, 3827–3844.